\documentclass{article}
\usepackage[utf8]{inputenc}
\usepackage{changepage}
\usepackage{graphicx}
\usepackage{subfiles}
\usepackage{amsmath}
\usepackage{hyperref}
\usepackage{longtable}
\usepackage[a4paper, margin=1in]{geometry}
\usepackage{booktabs}
\usepackage{authblk}
\usepackage{cleveref}
\usepackage{caption}
\usepackage{ifthen}
\newboolean{designheaders}
\usepackage{bbding}
\usepackage[backend=biber, style=numeric, sorting=none, maxnames=6, minnames=3, giveninits=true]{biblatex}
\usepackage{tabularx}
\usepackage{pdflscape}
\usepackage{lineno}
\usepackage{etoolbox}
\usepackage{fontawesome5}
\usepackage{array}
\newcolumntype{R}[1]{>{\raggedleft\arraybackslash}p{#1}}

\renewbibmacro*{journal+issuetitle}{
  \printfield{journaltitle}
  \setunit*{\addspace}
  \printfield{year}
  \printfield{volume}
  \newunit
}
\renewbibmacro{in:}{}
\DeclareFieldFormat[article]{title}{#1}
\DeclareFieldFormat[inproceedings]{title}{#1}

\title{Designing UNICORN: a Unified Benchmark for Imaging in Computational Pathology, Radiology, and Natural Language}

\author[a,*]{Michelle Stegeman}
\author[b,*]{Lena Philipp}
\author[b,*]{Fennie van der Graaf}
\author[a,*]{Marina D'Amato}
\author[a,c,*]{Cl\'ement Grisi}
\author[b,*]{Luc Builtjes}
\author[b,*]{Joeran S. Bosma}
\author[a,c,*]{Judith Lefkes}
\author[b,*]{Rianne A. Weber}
\author[b]{James A. Meakin}
\author[d,e]{Thomas Koopman}
\author[b]{Anne Mickan}
\author[b]{Mathias Prokop}
\author[b]{Ewoud J. Smit}
\author[a]{Frédérique Meeuwsen}
\author[a,c]{Geert Litjens}
\author[a]{Jeroen van der Laak}
\author[b]{Bram van Ginneken}
\author[b]{Maarten de Rooij}
\author[b]{Henkjan Huisman}
\author[b]{Colin Jacobs}
\author[a,$\dagger$,\Envelope]{Francesco Ciompi}
\author[b,$\dagger$,\Envelope]{Alessa Hering}
\author[$\diamond$]{on behalf of the UNICORN consortium}

\affil[a]{Department of Pathology, Radboud University Medical Center, Nijmegen, The Netherlands}
\affil[b]{Department of Medical Imaging, Radboud University Medical Center, Nijmegen, The Netherlands}
\affil[c]{Oncode Institute, Utrecht, the Netherlands}
\affil[d]{Informatics Institute, Faculty of Science, University of Amsterdam, Amsterdam, The Netherlands}
\affil[e]{Departments of Biomedical Engineering and Physics, University Medical Center Amsterdam, Amsterdam, The Netherlands}
\affil[*]{These authors contributed equally to this work and may be listed in any order.}
\affil[$\dagger$]{These authors contributed equally to this work.}
\affil[$\diamond$]{The consortium author lists are provided in the Consortium Authors section.}
\affil[\Envelope]{Corresponding authors: \href{mailto:Francesco.Ciompi@radboudumc.nl}{Francesco.Ciompi@radboudumc.nl}
\href{mailto:Alessa.Hering@radboudumc.nl}{Alessa.Hering@radboudumc.nl}}

\date{}

\begin{document}
\maketitle

\begin{abstract}
Foundation models are changing the way we develop medical artificial intelligence. By learning broadly generalizable features across diverse data modalities, a single model can be rapidly adapted to address multiple modalities and tasks with minimal supervision. This potential comes with the urgent need to reliably benchmark, understand and compare the performance and clinical impact of foundation models across data modalities and clinical tasks.  We introduce UNICORN, a fundamentally new benchmarking concept for medical foundation models. UNICORN brings four main contributions to medical artificial intelligence. First, a framework that enables a \emph{one-to-many} benchmarking approach, where a single foundation model is tested across multiple tasks and data modalities. Here, we populate it with 20 tasks across radiology, pathology, and clinical text, covering classification, detection, segmentation, regression, and vision–language generation. Second, a publicly available evaluation platform that implements, for the first time, a two-step approach to run foundation models for data encoding followed by custom task-specific adaptation via few-shot learning and linear probing mechanisms. Third, we create a meta-model that combines state-of-the-art foundation models in pathology, radiology and language with novel task-specific adapters that address all UNICORN tasks, which we refer to as \emph{Unicorn Model-0} (UM-0). Finally, we design a novel UNICORN score to benchmark and compare model performance across all tasks. We present the results of UM-0 using sequestered test data from over 2,400 patients, 3,700 vision cases, and 2,400 clinical reports from 17 institutions across eight countries, spanning eight anatomical regions and four imaging modalities.  Data, baselines, and evaluation platform are publicly accessible at \url{unicorn.grand-challenge.org}.
\end{abstract}

\begin{refsection}
% INTRODUCTION
\section{Introduction}
The application of artificial intelligence (AI) to medicine is increasingly delivering technology with potential for being adopted into diagnostic workflows \cite{leeuwen_clinical_use}. This progress goes hand in hand with growing needs for standardized evaluation frameworks that enable transparent and reproducible comparison of models across diseases, imaging domains, and clinical settings.  Public challenges have \emph{de-facto} become a reference tool to benchmark medical AI models. Since their inception, challenges have followed a ``one-to-one'' design, where one algorithm competes on a single, narrowly defined task, typically restricted to a single anatomical region, data modality, or output target. Examples of such challenges include PI-CAI \cite{saha2024artificial} for tumor detection in prostate MRI, TIGER \cite{tiger} for quantification of tumor-infiltrating lymphocytes in breast cancer digital pathology, and LUNA16 \cite{SETIO2017} for the detection of pulmonary nodules in CT scans, among others. 

In recent years, the scientific community has witnessed a fast-growing release of medical Foundation Models (FMs). Based on data encoders pre-trained using large-scale medical data spanning multiple domains, modalities, anatomical regions, and disease contexts, typically using self- or weakly supervised objectives \cite{chen2020simple, brown2020language, devlin2019bert}, FMs produce rich, general purpose data representations that capture generic features across diverse inputs. Models like CTransPath \cite{wang2022transformer}, Phikon \cite{filiot2023scaling}, and UNI \cite{chen2024towards} were the first examples of image encoders in pathology, followed by CT-FM \cite{pai2025vision} and Foundation Model for Cancer-Imaging Biomarkers \cite{pai2024foundation} for CT radiology, and USFM \cite{jiao2024usfm} for ultrasound. In parallel, developments in vision-language models (VLMs) delivered models such as MedCLIP \cite{wang2022medclip}, Med-Gemini \cite{saab2024capabilities}, and Med-PaLM \cite{tu2024towards}, along with domain-focused variants such as CT-CLIP \cite{hamamci2024foundation} and Merlin \cite{blankemeier2024merlin} for CT, ELIXR \cite{xu2023elixr} and BioViL-T \cite{bannur2023learning} for X-ray, and Quilt-1M \cite{ikezogwo2023quilt} and CONCH \cite{lu2024visual} for pathology.

A common characteristic of these models is that they produce data representations that can then be adapted to new downstream tasks with minimal additional training, usually via \emph{few-shot} learning techniques, reducing annotation needs and improving robustness to variable conditions often present in clinical practice, such as different image scanners, institutions, and populations. 
In this context, we need to shift performance evaluation from one-to-one to ``one-to-many'', to measure generalization of FMs across diverse domains, modalities, and tasks.

Some benchmarking initiatives enabled systematic, broad, multi-organ, and multimodal evaluations.  Examples are the Medical Segmentation Decathlon \cite{antonelli2022medical}, FLARE \cite{MedIA-FLARE21}, AMOS \cite{ji2022amos}, CHAOS \cite{CHAOS2021}, Learn2Reg \cite{hering2022learn2reg} and MOOD \cite{zimmerer2022mood}. However, their design does not support benchmarking via few-shot learning and are often limited to a narrow, specific task, such as segmentation or registration only.

Recent benchmarks have demonstrated the value of systematic evaluation of FMs, including the works of \cite{campanella2025clinical}, \cite{wang2023real},  \cite{kaiko.ai2024eva} and \cite{jin2024fairmedfm}. However, existing benchmarks differ in their design. Evaluation is often narrow in scope, restricted to one task-type, such as classification; to a single modality, such as pathology; or organized around a specific aspects such as fairness. Moreover, test data is either not publicly accessible, limiting reproducibility and broad comparison of current and future models, or already public, risking pretraining leakage and potentially undermining the validity of the reported performance.

The recent DRAGON benchmark \cite{bosma2025dragon} addresses this issue with a systematic, transparent, and comparable assessment of model capabilities via indirect access to sequestered data, demonstrating a promising approach to maintain benchmark integrity and accessibility. However, this approach is limited to medical LLMs, in line with initiatives such as HealthBench \cite{arora2025healthbench} and MedHELM \cite{bedi2025medhelm}. 

% FIGURE 1
\begin{figure}
    \centering
    \includegraphics[width=0.5\textwidth]{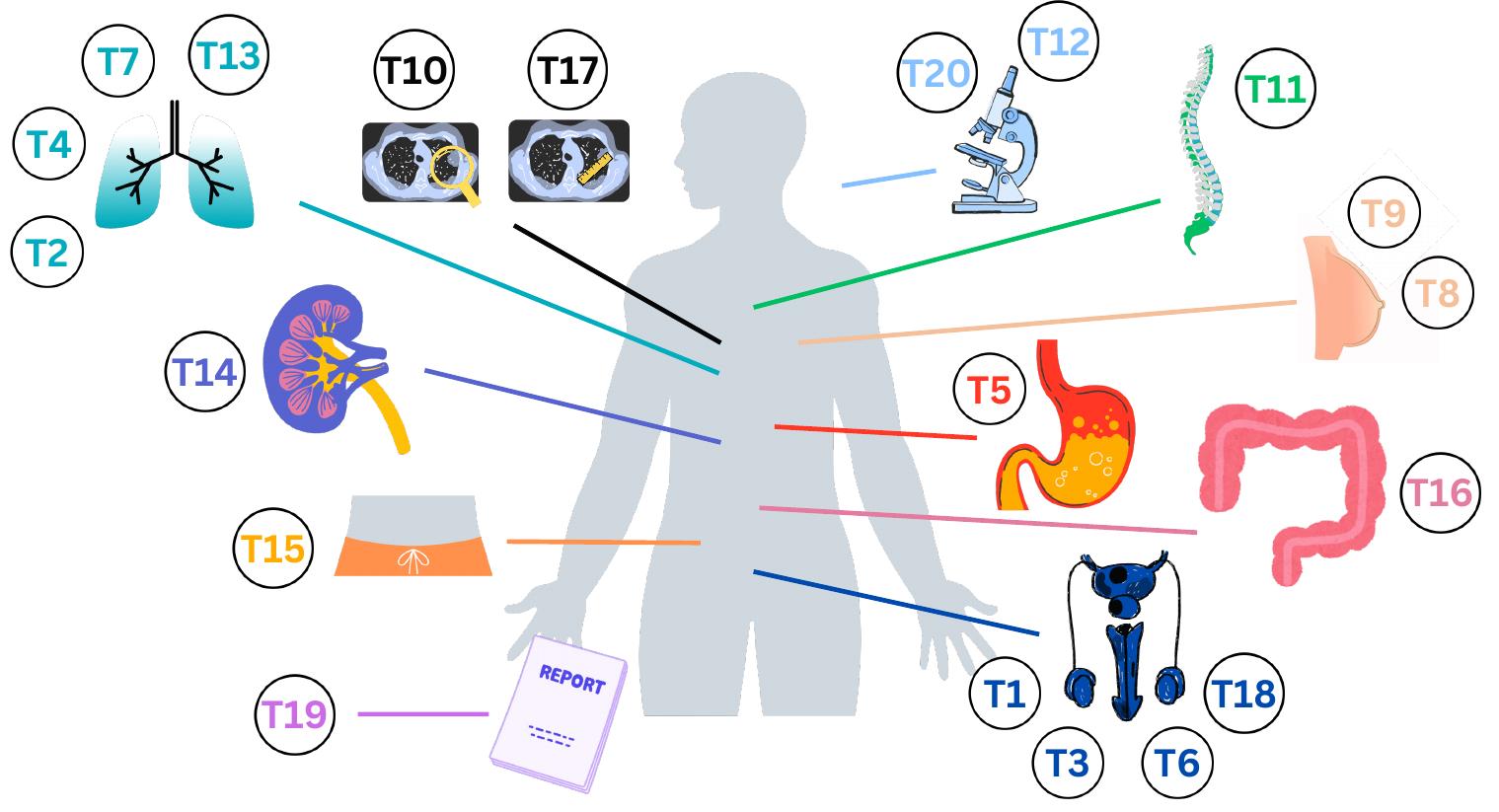}
    \caption{\textbf{Overview of the 20 UNICORN benchmark tasks.} The benchmark includes 20 tasks, of which 15 focused on eight different anatomic regions, whereas the remaining five are broad-scope tasks that span multiple regions or represent specific medical processes.}
    \label{fig:bodyparts}
\end{figure}

% FIGURE 2
\begin{figure*}[t]
    \centering
    \includegraphics[width=\textwidth]{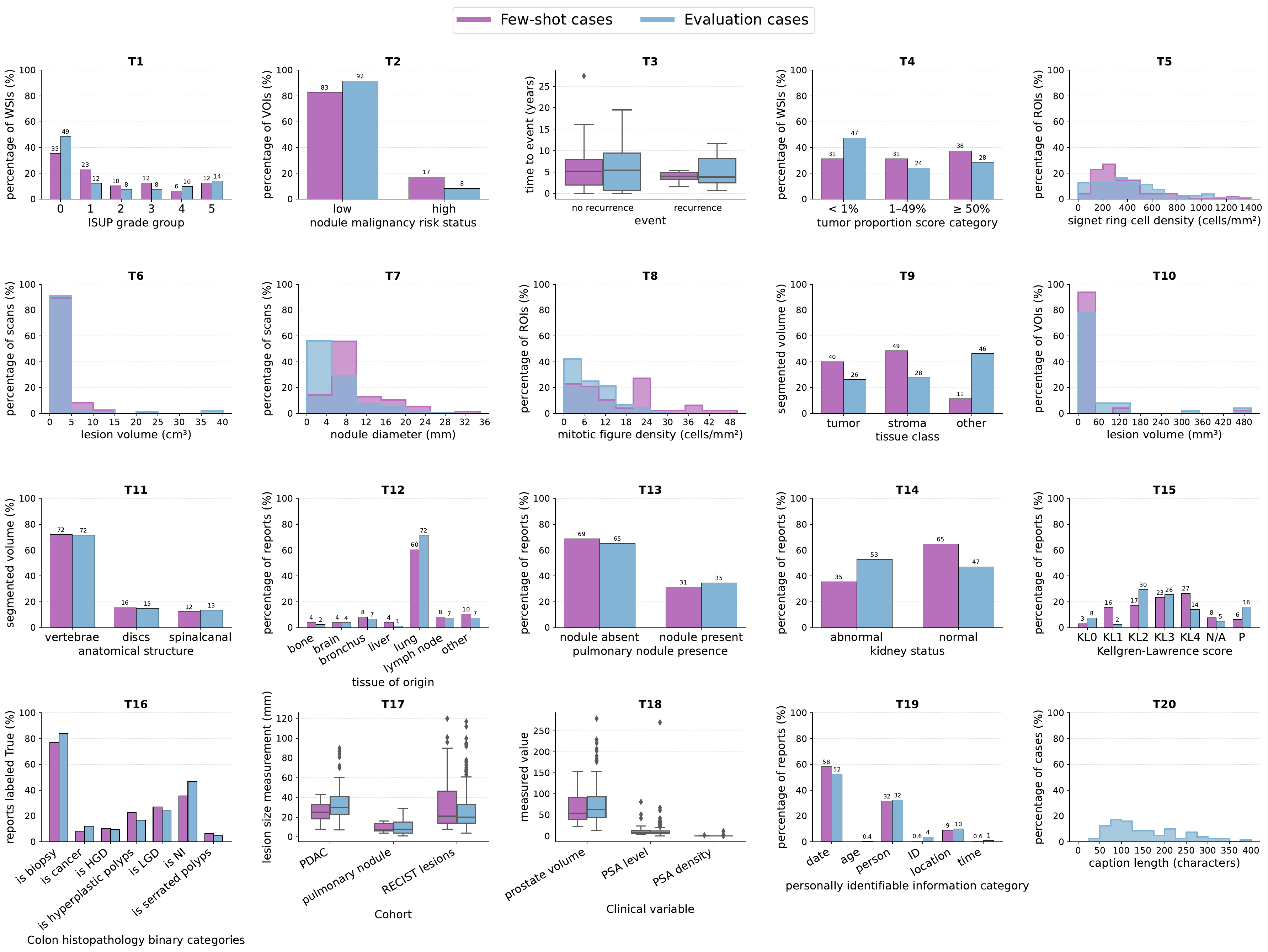}
    \caption{\textbf{Reference labels in the UNICORN validation set.} For classification tasks, \textbf{T1, T2, T4, T12-T16}, bar charts show the proportion of cases per label. For \textbf{T16}, which contains seven independent binary labels, each bar represents the proportion of reports labeled True, with the remainder corresponding to the proportion of False labels. Regression tasks, \textbf{T3, T17, T18}, are summarized with boxplots of target values, detection tasks, \textbf{T5-T8}, with histograms of object counts, and segmentation tasks, \textbf{T9-T11}, with bar charts showing either the proportion of each class for multiclass tasks, or the proportion of object volume for single class tasks. For named entity recognition, \textbf{T19}, bars show the distribution of target categories, and for caption generation, \textbf{T20}, the distribution of caption lengths is shown, few-shots are not available for this task.}
    \label{fig:validation_distribution}
\end{figure*}

We introduce a Unified beNchmark for Imaging in COmputational pathology, Radiology, and Natural language (UNICORN) as  the first multi-tasks, multi-domain, and multi-modality open benchmark for medical foundation models in a single, standardized framework. With UNICORN, we bring a fundamentally new benchmarking concept for medical foundation models, consisting in four novel contributions that bridge the gaps of current benchmarks in medical AI while setting the basis for present and future analysis of medical FMs.

First, a framework that enables a \emph{one-to-many} benchmarking approach, where a single foundation model is tested across multiple tasks and data modalities. We integrated 20 tasks spanning radiology, digital pathology, clinical language understanding, and vision-language generation. All tasks use sequestered test sets derived from real-world clinical cohorts to ensure unbiased and reproducible evaluation. Several tasks are derived from projects that build upon established challenges, which include clinically validated reference standards and enable comparison with task-specific models.

Second, we deliver a publicly available evaluation platform that implements, for the first time, a two-step approach to use foundation models for data encoding followed by custom task-specific adaptation via few-shot learning and linear probing. This new framework decouples foundation-model inference from lightweight task-specific adaptation, isolating the generality and quality of learned representations, and provides a consistent and reproducible basis for comparing heterogeneous models across tasks and modalities while reflecting constraints of clinical data curation, where labeled data are scarce.

Third, we create and release a meta-model based on state-of-the-art foundation models in pathology, radiology and language with novel task-specific adapters that address all UNICORN tasks, which we refer to as \emph{Unicorn Model-0} (UM-0).

Finally, we design a novel UNICORN score to benchmark and compare model performance across all tasks using sequestered test data from over 2,400 patients, 3,700 vision cases, and 2,400 clinical reports from 17 institutions across eight countries, spanning eight anatomical regions and four imaging modalities. 
In this paper, we describe the design of UNICORN, its data and evaluation process, the design of UM-0 and report on its results on the full benchmark, providing the first evidence of model generalization under the unified protocol.

% DATA
\section{Data and tasks}
UNICORN introduces a diverse set of 20 clinically motivated tasks, spanning radiology, pathology, and natural language, to assess whether a single pre-trained model can support heterogeneous clinical objectives using a shared representation. For this purpose, included data spans multiple 1) data sources, 2) anatomical regions, 3) task types, and 4) clinical domains. 

In terms of \emph{data sources}, datasets were obtained from 17 institutions across eight countries and were collected from sequestered data of real-world clinical cohorts. All data were fully anonymized before inclusion. In several cohorts, data were previously used in challenges or clinical projects, where ethical approval or consent waivers were obtained. Table \ref{tab:tasks_overview} provides an overview of all tasks, including task type, modality, domain, and evaluation metrics. Detailed task descriptions are available in \ref{app:tasks}.

In terms of  \emph{anatomical regions}, among the introduced 20 tasks, 15 of them are specific to anatomical regions, while the remaining five are broad-scope tasks spanning multiple anatomical regions, or representing general clinical processes (Figure \ref{fig:bodyparts}). Across the 15 region-specific tasks, eight anatomical regions are represented: lung, hip, kidney, prostate, colon, stomach, breast, and spine. 

In terms of  \emph{task types}, across 20 tasks, UNICORN includes eight classification tasks, four detection tasks, three regression tasks, three segmentation tasks, one named entity recognition task, and one caption generation task. This diversity requires algorithms to handle both global semantic reasoning and spatially grounded predictions, as well as language understanding. Of the 20 tasks, 11 are vision-only tasks, eight are language-only tasks, and one is a vision-language task. 

In terms of \emph{clinical domains}, UNICORN covers 10 radiology tasks, nine pathology tasks, and one joint task spanning radiology and pathology. This balanced distribution allows for systematic evaluation of whether a foundation model learns representations that transfer across domains and modalities under a unified protocol. 

UNICORN intentionally reflects the heterogeneous and often imbalanced nature of real-world clinical data. Figure \ref{fig:validation_distribution} summarizes label distributions across all 20 tasks in the UNICORN validation set, showing substantial variation in class prevalence, outcome frequencies, and annotation density. For tasks that support few-shot adaptation, distributions are shown separately for few-shot examples and evaluation case data, illustrating the degree to which the examples reflect the tasks' evaluation cases. 

By design, all UNICORN data is sequestered and therefore only indirectly accessible. However, to support model developers and facilitate their submission processes, we identified a set of publicly available data related to several UNICORN tasks, which may be used for model pretraining or local experimentation prior to model benchmarking on the UNICORN platform. Table \ref{app_tab:dataset_info} provides an overview of these public resources, including usage license, public links, and associated publications.

% FIGURE 3
\begin{figure}[h]
    \centering
    \includegraphics[width=\linewidth]{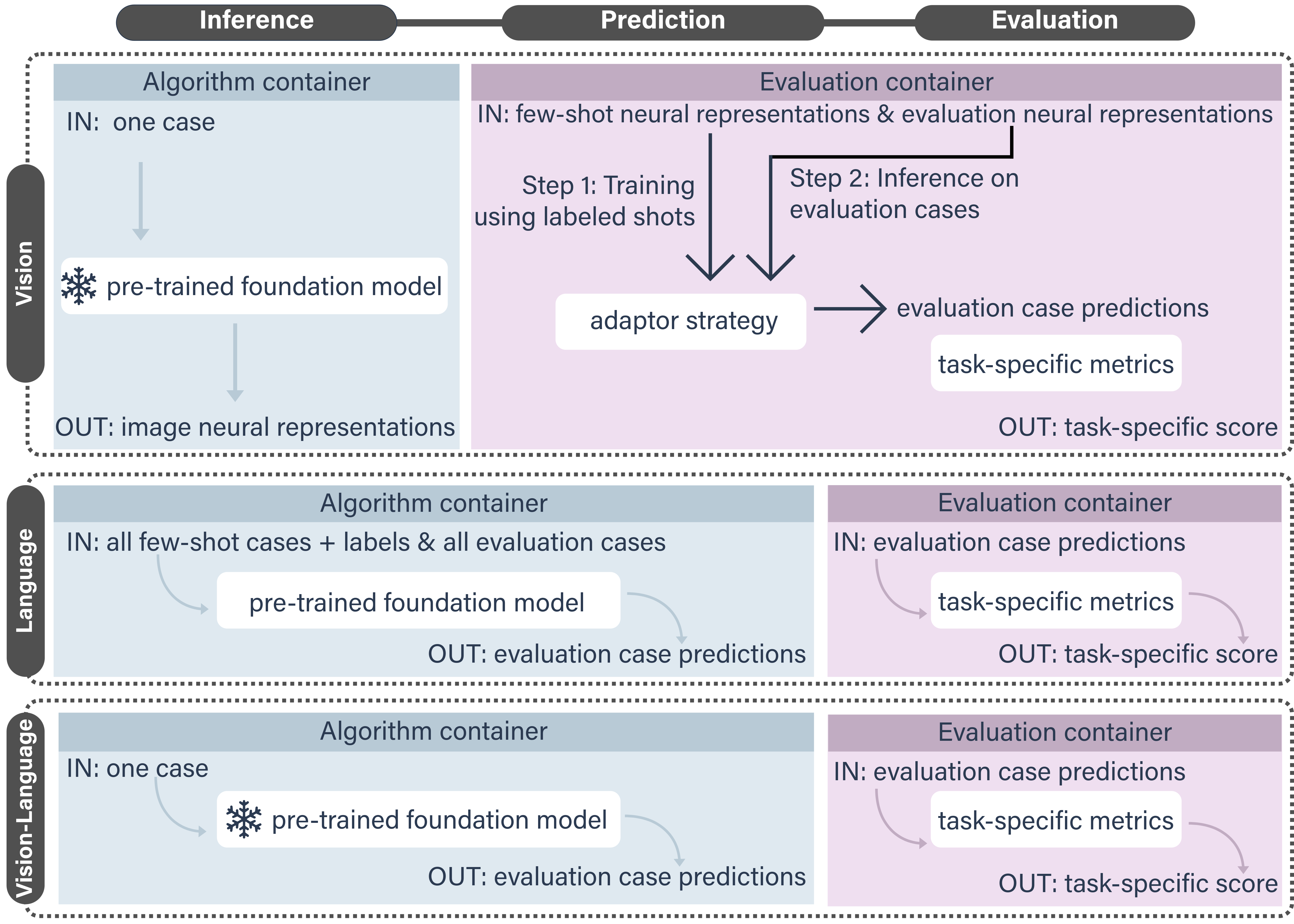} 
    \caption{\textbf{UNICORN benchmarking pipeline} The pipeline is structured vertically from data storage to metrics reporting, with modality-specific differences shown horizontally for vision, language, and vision--language.
    \textit{Vision tasks:} Each case is processed by the Algorithm container to extract generic representations using pre-trained foundation models. These representations are passed to the Evaluation container, where a lightweight adaptor trained on labeled few-shot examples produces task-specific predictions and computes evaluation metrics. 
    \textit{Language tasks:} All labeled few-shot cases and evaluation cases are together provided to the Algorithm container, which generates predictions that are evaluated in the Evaluation container.
    \textit{Vision-language task:} Each case is processed individually by the Algorithm container, which uses the textual task description to generate a textual prediction that is evaluated in the Evaluation container.
    For all tasks, the resulting metrics are reported on their respective leaderboards on Grand Challenge.
    }
    \label{fig:unicorn_pipeline}
\end{figure}

\newcommand{\casescell}[3]{
  \makebox[0.5cm][r]{#1} / 
  \makebox[0.5cm][r]{#2} / 
  \makebox[0.5cm][r]{#3}
}

\begin{table*}[!h]
\centering
\footnotesize
\caption{Overview of the tasks in the UNICORN challenge. \textit{AUROC = area under the receiver operating characteristic curve, CLS = classification, DET = detection, GEN = caption generation, H\&E = Hematoxylin and Eosin, IHC = immunohistochemistry, ISUP = International Society of Urological Pathology, MIXED = radiology and pathology, NER = named entity recognition, NSCLC = non-small cell lung cancer, PATH = Pathology, PSA = prostate-specific antigen, RAD = Radiology, REG = regression, ROIs = regions of interest, SEG = segmentation, WSI = whole-slide image.}}
\label{tab:tasks_overview}
\begin{tabular}{l p{3.5cm} l l p{2.8cm} p{2.6cm} p{1.6cm}}
\toprule
\textbf{ID} & \textbf{Name} & \textbf{Type} & \textbf{Modality} & \textbf{Metric} & \textbf{Cases (few-shot / validation / test)} & \textbf{Time limit (min)} \\
\midrule
\multicolumn{7}{@{}l@{}}{\normalsize\textbf{Vision tasks}} \\
\addlinespace[0.3em]
T1  & ISUP scoring in H\&E prostate biopsies & CLS & PATH & Quadratic weighted kappa & \casescell{48}{195}{113} & 10 \\
T2  & Lung nodule malignancy in CT & CLS & RAD & AUROC & \casescell{64}{108}{533} & 5 \\
T3  & Time to biochemical recurrence in H\&E prostatectomies & REG & PATH & Censored c-index & \casescell{48}{49}{521} & 25 \\
T4  & Tumor proportion score in NSCLC IHC WSI & CLS & PATH & Quadratic weighted kappa & \casescell{48}{116}{474} & 10 \\
T5  & Signet ring cells in H\&E ROIs of gastric cancer & DET & PATH & F1 score & \casescell{48}{79}{348} & 10 \\
T6  & Clinically significant prostate cancer in MRI & DET & RAD & Average of AUROC and AP & \casescell{48}{100}{400
}& 10 \\
T7  & Lung nodule detection in thoracic CT & DET & RAD & Sensitivity & \casescell{48}{83}{83} & 5 \\
T8  & Mitotic figures in breast cancer H\&E ROIs & DET & PATH & F1 score & \casescell{48}{180}{400} & 10 \\
T9  & Tumor and stroma segmentation in breast H\&E & SEG & PATH & Dice & \casescell{48}{24}{33} & 5 \\
T10 & Universal lesion segmentation in CT ROIs & SEG & RAD & Dice, long- and short-axis errors & \casescell{48}{50}{725} & 10 \\
T11 & Anatomical segmentation in lumbar spine MRI & SEG & RAD & Dice & \casescell{48}{48}{97} & 10 \\
\midrule 
\multicolumn{7}{@{}l@{}}{\normalsize\textbf{Language tasks}} \\
\addlinespace[0.3em]
T12 & Histopathology sample origin & CLS & PATH & Unweighted kappa & \casescell{48}{215}{297}  & 240/240\textsuperscript{*}  \\
T13 & Pulmonary nodule presence & CLS & RAD & AUROC & \casescell{48}{300}{200} & 120/240\textsuperscript{*} \\
T14 & Kidney abnormality & CLS & RAD & AUROC & \casescell{48}{125}{183} & 120/240\textsuperscript{*} \\
T15 & Hip Kellgren-Lawrence scoring & CLS & RAD & Unweighted kappa & \casescell{32}{100}{108} & 120/240\textsuperscript{*} \\
T16 & Colon histopathology diagnosis & CLS & PATH & Macro AUROC & \casescell{48}{250}{500} & 120/240\textsuperscript{*} \\
T17 & Lesion size measurements & REG & RAD & RSMAPE & \casescell{48}{242}{298} & 120/240\textsuperscript{*} \\
T18 & Prostate volume and PSA (density) & REG & RAD & RSMAPE & \casescell{48}{250}{500} & 120/240\textsuperscript{*} \\
T19 & Report anonymization & NER & MIXED & Weighted F1 & \casescell{48}{200}{400} & 120/240\textsuperscript{*} \\
\midrule
\multicolumn{7}{@{}l@{}}{\normalsize\textbf{Vision-Language task}} \\
\addlinespace[0.3em]
T20 & WSI captioning & GEN & PATH & BLEU-4, ROUGE-L, METEOR, CIDER, BERTscore & \casescell{0}{81}{310} & 25 \\
\bottomrule
\end{tabular}
\begin{flushleft}
\footnotesize \textsuperscript{*} Time limit given as validation/test.
\end{flushleft}
\end{table*}

% METHODS
\section{Methods}
In this section, we describe 1) the main components of the UNICORN platform, the adopted approaches for data storage, task definition across vision, language and vision-language fields, and model execution, which are necessary to understand model submission and evaluation processes; 2) the design and implementation of the proposed UM-0 meta-encoder and of the multiple adaptors made available via UNICORN; 3) the evaluation process, including the details of the novel UNICORN score that we introduced.

% UNICORN platform
\subsection{UNICORN platform}
The UNICORN benchmark requires a web-based, publicly accessible platform that manages secure data storage, model execution and evaluation. For this purpose, we built UNICORN on the Grand Challenge (GC) platform, developed by \cite{meakin_grand-challengeorg_2023} \footnote{\url{grand-challenge.org}}, which addresses all the aforementioned needs.

% two-step procedure
\paragraph{Two-step procedure for few-shot learning}
Based on GC, we developed a novel \emph{two-step} architecture that separates foundation model inference from task-specific evaluation, within a unified pipeline that spans data storage, algorithm inference, task-specific predictions, task-specific metric computation, and leaderboard reporting. The first step focuses on the \textit{Algorithm}, here defined as a single or multiple pre-trained foundation models, containing the entire collection of all dependencies and code required for model execution. On GC, the Algorithm is packaged using modern container image technology and is uploaded to the platform as single unit. 

The second step focuses on the \textit{Evaluation}, which consists in a) performing task-specific adaptation, and 2) computing task-specific evaluation metrics. The UNICORN organizers fully maintain the evaluation part. Both task-specific adaptations (i.e., adaptors) and evaluation metrics are pre-defined, and adaptors can be expanded and modified upon request. In practice, this is implemented via GitHub pull requests.

% adaptation via few-shot learning
The proposed algorithm-evaluation structure implements a benchmark based on few-shot learning. On the one hand, the role of the algorithm is in line with common data encoding steps performed by pre-trained foundation models on large archives. On the other hand, the adaptors present in the evaluation step have access to additional, smaller, datasets of labelled few-shot input samples. These can be used by the adaptors, on-the-fly, to build adaptor-specific components trained exclusively using the labeled few-shot cases for each task. In this mechanism, no pre-trained weights are permitted within the adaptors themselves, which in practice are all initialized from scratch and optimized using the available few-shot data.

% algorithms run on data
Submitted algorithms run on datasets that are organized in sequestered (i.e., not directly accessible, not downloadable, not viewable) archives on GC. For efficiency, each archive item contains a single case for vision and vision-language tasks, or a combined set of reports for language tasks. Depending on the task, a case may represent a patient examination, a whole-slide image, or a specific region of interest (see \ref{app:tasks} for task definitions). For vision and vision–language tasks, cases are provided individually to the Algorithm container without revealing whether they belong to the few-shot or evaluation set. For language tasks, all cases, including both the labeled few-shot examples and the evaluation cases requiring prediction, are provided to the algorithm at once.

% Algorithm Inference
The Algorithm container autonomously processes each case provided by the platform. For vision tasks, it extracts generic representations suitable for downstream adaptation, whereas for language and vision-language tasks, it generates task-specific predictions, which are passed unchanged to the Evaluation container. All required steps, such as preprocessing, prompting, and language-specific tuning steps, are executed within the container without internet access to ensure data security. Each archive item is accompanied by a task configuration file specifying domain, modality, task type, and output shape requirements. Publicly available templates and baseline resources relevant to this workflow are summarized in \ref{app:example_code}.

\subsection{Task-specific implementation}
% vision
For \emph{vision} tasks, the form of the extracted representation depends on the prediction granularity. Dense prediction tasks, such as segmentation and detection, multiple patch-level representations may be extracted to capture spatially resolved information. In contrast, case-level tasks, such as classification and regression, require aggregation into a single representation per case, thereby aligning the Algorithm output with the granularity of the reference labels.

% vision-language
For \emph{vision-language}, prediction may be guided by the task description provided configuration, and postprocessing can include translation into the target label language.

% language
For each \emph{language} task, the prediction pipeline may include preprocessing (e.g., text-normalization), the use of a pretrained foundation model, fine-tuning or prompting on the few-shot dataset, and postprocessing.

\begin{table*}[ht]
    \small
    \caption{Overview of baseline models used across modalities and tasks.}
    \label{tab:baseline_models}
    \centering
    \begin{tabular}{llp{5.5cm}}
    \hline
    \textbf{Model} & \textbf{Modality} & \textbf{Tasks} \\
    \hline
    Virchow \cite{virchow} & Vision (pathology) & Detection, segmentation \\
    PRISM \cite{prism} & Vision-Language (pathology) & Caption generation, classification, regression \\
    MRSegmentator \cite{H_ntze_2025} & Vision (MR Radiology) & Detection, segmentation \\
    DINOv2-small \cite{oquab2023dinov2} & Vision (CT Radiology) & Classification \\
    CT-FM \cite{pai2025vision} & Vision (CT, Radiology) & Detection, Segmentation \\
    Phi-4 \cite{abdin2024phi} & Language & classification, regression, NER \\
    Opus-MT (EN--NL) & Language & Translation \\
    \hline
    \end{tabular}
\end{table*}

% UM-0
\subsection{UNICORN Model-0}
Due to the absence of a publicly available model that natively supports all modalities, we developed a baseline implementation based on a modular design that uses state-of-the-art approaches within each individual modality. We refer to this meta-model as Unicorn Model-0 (UM-0), which provides a comprehensive solution to address all vision, language, and vision-language tasks while maintaining a consistent and reproducible evaluation framework across data types. The specific models used in UM-0 are listed in Table~\ref{tab:baseline_models}.

For the vision tasks, UNICORN evaluates foundation models through lightweight task-specific adaptors that operate on frozen model embeddings. These adaptors are designed to convert generic foundation model representations into task-specific predictions while limiting the amount of trainable task-specific machinery. 

The adaptor strategy depends on the granularity of the task output. For case-level classification and regression tasks, the representation of each case is summarized into a single embedding. Classification adaptors for UM-0 cover linear probing for pathology vision classification, and a shallow multilayer perceptron for the radiology vision classification tasks. For survival prediction tasks, analogous shallow multilayer perceptrons with trainable regression heads perform continuous prediction using a discretized time-to-event formulation and a survival-aware objective.

For spatial prediction tasks, the adaptors operate on patch-level or spatially arranged embeddings. Detection tasks use lightweight decoders that map local embeddings to object probabilities or candidate locations. In pathology cell-detection tasks, patch-level embeddings are converted into low-resolution probability maps from which candidate cell locations are extracted. In radiology point-detection tasks, patch embeddings and their spatial metadata are used to predict candidate lesion locations in patient coordinates. Segmentation tasks use lightweight upsampling decoders that reconstruct dense pixel- or voxel-wise predictions from spatially arranged embeddings. For pathology, this produces two-dimensional tissue segmentations; for radiology, patch-level predictions are reassembled into a three-dimensional volume using the target image geometry.

All adaptor-specific parameters are initialized from scratch and trained exclusively on the few-shot examples provided for the corresponding task. No pretrained weights are allowed within the adaptors. Thus, performance reflects the utility of the submitted frozen representations when combined with a standardized, low-capacity adaptation protocol. Detailed architectural choices, hyperparameters, and postprocessing steps for each adaptor are provided in the Supplementary Methods and in the public evaluation repository\footnote{\url{https://github.com/DIAGNijmegen/unicorn_eval}}. Beyond the baseline configurations described here, the UNICORN evaluation framework supports additional adaptor strategies, ranging from non-parametric methods such as k-nearest neighbors to lightweight learning-based approaches, and can be extended with community-contributed adaptors through the official evaluation repository.

% Evaluation
\subsection{Evaluation}
The Evaluation container transforms the Algorithm outputs into task-specific performance scores and computes the metrics used for leaderboard ranking. For all tasks, predictions generated or derived from the Algorithm output are compared against sequestered reference labels to assess the task-specific performance (see Table \ref{tab:tasks_overview}). By centralizing all evaluation in a single container, UNICORN ensures reproducible and consistent assessment across all submissions, modalities, and domains. 

\paragraph{UNICORN Score}
\label{sec:UNICORN_score}
Evaluating the overall foundational characteristics of foundation models across multiple tasks and modalities requires a unified metric that summarizes overall performance. Since each task in UNICORN uses a different evaluation measure, ranging from Dice similarity to AUROC and mean absolute error, these values must be made comparable to assess how well a foundation model performs overall. The UNICORN Score addresses this need by combining task-specific metrics into a single, standardized measure of generalizability.

Performance on each downstream task was first computed using its task-specific metric (\ref{tab:tasks_overview}). To aggregate results across heterogeneous metrics and task types, all scores were normalized to a common 0--1 range using a consistent scaling scheme. For each task $n$, the normalized score $t_n$ was calculated as

\begin{equation}
t_n = \frac{S_n - S_n^{\mathrm{ref}}}{S_n^{\max} - S_n^{\mathrm{ref}}},
\label{eq:tn}
\end{equation}

where $S_n$ denotes the raw task-specific score, $S_n^{\mathrm{ref}}$ denotes the performance of a minimal reference model (e.g., a majority-class predictor for classification tasks), and $S_n^{\max}$ represents the maximum achievable value of the corresponding metric. This normalization ensures all metrics are on a common scale. Using a minimal reference model anchors normalization at a meaningful lower bound, ensuring that scores reflect improvement over trivial solutions.

The overall UNICORN Score, $S_\mathrm{UNICORN}$, is defined as the mean of the normalized scores across all $N = 20$ tasks:

\begin{equation}
S_{\mathrm{UNICORN}} = \frac{1}{N} \sum_{n=1}^{N} t_n.
\label{eq:unicorn_score_all}
\end{equation}

All tasks were weighted equally to reflect a uniform, cross-task generalization, rather than prioritizing specific domains or dataset sizes. This aggregated score provides an interpretable measure of how well a foundation model generalizes across diverse tasks and modalities. Domain-specific UNICORN Scores (for pathology, radiology, and language) were computed analogously by averaging normalized scores within each domain. Task-specific normalization constants are provided in Eq. \ref{app_eq:unicorn_score}.

\paragraph{UNICORN Phases}
UNICORN consists of four sequential phases. First, public example datasets \cite{damato_unicorn_2025} were released during the off-platform development phase to support local prototyping and to ensure compatibility with the container-based execution used in later stages. Second, a non-scored check phase used a small, carefully selected dataset containing edge cases, e.g., the largest image from the cohort, to verify that each user's Algorithm container image could execute end-to-end within time on the Grand Challenge platform. Successful completion of the check phase was required for entry into subsequent phases. Third, the validation phase provided users with intermediate performance feedback on sequestered data. To support detailed evaluation, task-specific leaderboards were made available, alongside three domain/ modality-specific combined leaderboards (pathology--vision, radiology--vision, and language), and an all-tasks leaderboard covering all 20 tasks. The all-tasks leaderboard is the primary endpoint, while the domain/ modality-specific leaderboards enable evaluation of more specialized algorithms. During validation, each user was limited to a total of six successful submissions per task: three to each task-specific leaderboard, two per combined leaderboard, and one to the all-tasks leaderboard. 

Finally, the test phase consisted of a single opportunity on sequestered data, either to the all-tasks leaderboard or to each of the combined leaderboards. Additional information about phases is provided in \ref{app:challenge_phases}.

\section{Experimental results}

\begin{table*}[]
\centering
\small
\caption{Baseline performance across the 20 UNICORN tasks. For each task, the primary evaluation metric, test set size (\textit{n}), raw performance, and normalized score are reported. Tasks are grouped by modality and task type.
}
\begin{tabular}{l p{4.0cm} l r c c}
\toprule
\textbf{ID} & \textbf{Name} & \textbf{Metric} & \textbf{Test \textit{n}} & \textbf{Raw} & \textbf{Normalized} \\
\midrule
\multicolumn{6}{@{}l@{}}{\normalsize\textbf{Pathology vision}} \\
\addlinespace[0.2em]
\multicolumn{6}{@{}l@{}}{\textit{Classification}} \\
T1 & ISUP scoring & Quadratic weighted kappa & 113 & 0.77 & 0.77 \\
T4 & Tumor proportion score & Quadratic weighted kappa & 474 & 0.46 & 0.46 \\
\addlinespace[0.2em]
\multicolumn{6}{@{}l@{}}{\textit{Regression}} \\
T3 & Time to biochemical recurrence & Censored c-index & 521 & 0.59 & 0.18 \\
\addlinespace[0.2em]
\multicolumn{6}{@{}l@{}}{\textit{Detection}} \\
T5 & Signet ring cells & F1 score & 348 & 0.24 & 0.24 \\
T8 & Mitotic figures & F1 score & 400 & 0.00 & 0.00 \\
\addlinespace[0.2em]
\multicolumn{6}{@{}l@{}}{\textit{Segmentation}} \\
T9 & Tumor and stroma segmentation & Dice & 33 & 0.52 & 0.36 \\
\midrule
\multicolumn{6}{@{}l@{}}{\normalsize\textbf{Radiology vision}} \\
\addlinespace[0.2em]
\multicolumn{6}{@{}l@{}}{\textit{Classification}} \\
T2 & Lung nodule malignancy & AUROC & 533 & 0.67 & 0.33 \\
\addlinespace[0.2em]
\multicolumn{6}{@{}l@{}}{\textit{Detection}} \\
T6 & Clinically significant prostate cancer & Average of AUROC
and AP & 400 & 0.25 & 0.01 \\
T7 & Lung nodule detection & Sensitivity & 83 & 0.00 & 0.00 \\
\addlinespace[0.2em]
\multicolumn{6}{@{}l@{}}{\textit{Segmentation}} \\
T10 & Universal lesion segmentation & Dice, long- and short-
axis errors & 725 & 0.04 & 0.04 \\
T11 & Anatomical segmentation, spine & Dice & 97 & 0.01 & 0.01 \\
\midrule
\multicolumn{6}{@{}l@{}}{\normalsize\textbf{Language}} \\
\addlinespace[0.2em]
\multicolumn{6}{@{}l@{}}{\textit{Classification}} \\
T12 & Histopathology sample origin & Unweighted kappa & 297 & 0.64 & 0.64 \\
T13 & Pulmonary nodule presence & AUROC & 200 & 0.86 & 0.73 \\
T14 & Kidney abnormality & AUROC & 183 & 0.91 & 0.82 \\
T15 & Hip Kellgren-Lawrence scoring & Unweighted kappa & 108 & 0.29 & 0.29 \\
T16 & Colon histopathology diagnosis & Macro AUROC & 500 & 0.87 & 0.74 \\
\addlinespace[0.2em]
\multicolumn{6}{@{}l@{}}{\textit{Regression}} \\
T17 & Lesion size measurements & RSMAPE & 298 & 0.90 & 0.60 \\
T18 & Prostate volume and PSA & RSMAPE & 500 & 0.94 & 0.74 \\
\addlinespace[0.2em]
\multicolumn{6}{@{}l@{}}{\textit{Named entity recognition}} \\
T19 & Report anonymization & Weighted F1 & 400 & 0.48 & 0.48 \\
\midrule
\multicolumn{6}{@{}l@{}}{\normalsize\textbf{Vision-language}} \\
\addlinespace[0.2em]
\multicolumn{6}{@{}l@{}}{\textit{Caption generation}} \\
T20 & WSI captioning & \shortstack{BLEU-4, ROUGE-L, METEOR, \\ CIDER, BERTscore} & 310 & 0.14 & 0.14 \\
\midrule
\multicolumn{4}{@{}l}{\textbf{UNICORN Score}} & & \textbf{0.38} \\
\bottomrule
\label{tab:baseline_scores}
\end{tabular}
\end{table*}

We evaluate the UNICORN benchmark implementation to address three central questions motivated by the need for standardized, reproducible evaluation of medical foundation models across domains: 1. Does the proposed framework enable end-to-end evaluation across heterogeneous medical tasks? 2. Does the baseline establish an informative reference point by producing non-trivial, non-ceiling, and differentiated performance across the UNICORN tasks? and 3. How does baseline performance vary with modality and task type under the evaluation protocol? 

To address these questions, we evaluated the baseline implementation across all 20 UNICORN tasks. Per-task performance is reported in ~\autoref{tab:baseline_scores}, using the task-specific primary metric (Table~\ref{tab:tasks_overview}) and the corresponding normalized score (Eq.~\ref{eq:tn}). The normalized scores were aggregated into per-domain score (vision-pathology, vision-radiology and language), as well as into an overall UNICORN Score defined in Eq.~\ref{eq:unicorn_score_all}.

The framework supported end-to-end evaluation across all 20 tasks. The UM-0 successfully executed on every UNICORN task and produced valid outputs across radiology, pathology, clinical text, and vision-language domains. All tasks were completed within the per-case time limits, and the task-specific adaptors generated outputs matching the required evaluation formats for all tasks.

The UM-0 also provided a non-saturated and differentiated reference point across tasks, as reflected in the overall UNICORN Score of 0.38 with normalized scores ranging from 0.00 to 0.82. With 18 of the 20 tasks achieving scores above 0.00 while no task reached the metric ceiling. Seven tasks achieved scores of at least 0.60, whereas five scored 0.05 or below. Together, these results show that UM-0 performance was neither uniformly near the lower bound nor saturated at the upper bound, but instead demonstrated variation across tasks, providing the targeted basis for further development and benchmarking.

Baseline performance further varied across modality and task type. At the domain level, language tasks showed the highest mean normalized score (0.63), followed by pathology-vision (0.34) and radiology-vision (0.08). Within the language domain, five out of eight tasks exceeded 0.60, with scores ranging from 0.29 (hip Kellgren-Lawrence scoring, T15) to 0.82 (kidney abnormality, T14). The vision-language caption generation task (T20) achieved a normalized score of 0.14, placing its performance below all of the language tasks.

Within pathology-vision, performance was heterogeneous across task types. The two classification tasks achieved the highest performance, with ISUP scoring (T1) achieving the highest score among all vision tasks (0.77), followed by tumor proportion scoring (T4; 0.46). Segmentation of tumor and stroma (T9) achieved a moderate normalized score (0.36), whereas lower performance was observed for signet ring cell detection (T5; 0.24) and time-to-biochemical-recurrence prediction (T3; 0.18). Mitotic figure detection (T8) yielded the lowest score within the pathology-vision domain (0.00). 

Radiology-vision tasks consistently resulted in lower scores than pathology-vision and language tasks, with all results at or below 0.33. Lung nodule malignancy classification (T2; 0.33) scored highest, whereas the detection and segmentation task uniformly showed low performance, ranging between 0.00 and 0.04. 

Overall, the domain- and task-level results point to a consistent pattern across task types: classification tasks generally produced the strongest scores, and dense prediction task performed less well. Within the dense prediction tasks, performance is further split by domain, with pathology outperforming radiology (0.20 versus 0.02).

\label{subsec6}
\section{Discussion}
UNICORN has been developed to address the lack of publicly available benchmarks for evaluating medical foundation models across a broad range of tasks, organs, domains, and modalities. It implements a "one model–many tasks`` evaluation framework for comparing pretrained encoders via few-shot adaptation on 20 tasks spanning pathology vision, radiology vision, language, and vision–language. Task-wise results are complemented by the UNICORN Score, which summarizes normalized performance under a unified evaluation protocol. 

A central question was therefore whether this framework demonstrated useful cross-task evaluation, which we assessed through the UM-0 baseline. The introduced UM-0 produces non-trivial predictions across the majority of tasks and achieved an overall UNICORN Score of 0.38. This indicates that representations learned in pretraining can provide useful signal on diverse downstream tasks through minimal task-specific adaptation rather than extensive fine-tuning. The language tasks produced the strongest results, with an average of 0.63 and five of eight normalized scores at or above 0.60, exceeding ten out of eleven vision tasks whose average is 0.22. The framework is designed to enable comparison across modalities, although differences in the evaluation protocols can affect the absolute performance of the different modalities. For vision, the encoder is frozen and only a lightweight adaptor is trained. Each input is processed independently into an embedding and the adaptor is trained on these embeddings. The performance therefore reflects that of the pretrained representation under limited task-specific adaptation. In contrast, models submitted to language tasks may use prompting or fine-tuning, allowing them to jointly access the few-shot examples and evaluation inputs during inference, the score represents end-to-end model performance. As a result, while the scores are within the shared framework, differences in their absolute performance should be interpreted with respect to the modality-specific evaluation protocol applied. The benchmark nevertheless accommodates all three modalities under a shared framework while allowing for these protocol differences. 

The results revealed the extent to which a single representation transfers across heterogeneous downstream tasks. The tasks were deliberately chosen to retain real-world complexity, which is also reflected in the spread of per-task baseline scores (0.00-0.82). Many derive from existing public challenges with expert-annotated reference standards and demonstrated clinical demand. UNICORN spans a broad range of clinical input scales, from whole-slide images and complete CT and MRI volumes to individual regions of interest and free-text reports, retaining the complexity of each rather than reducing tasks to a uniform, simplified format. Tasks feature multi-sequence and multi-scanner acquisitions, large and variable input sizes, sparse or small targets within large fields, and substantial class imbalance. Individual tasks carry their own characteristic difficulty, such as the joint reasoning over three separately acquired and unregistered MR sequences required for clinically significant prostate cancer detection (T6), or the detection and precise localization of sparse mitotic figures, within regions of approximately 39 million pixels, requiring the localization of each target within a region of only about 900 pixels (T8). This matters for interpretation, as performance on UNICORN measures performance under conditions that approximate clinical use. Because tasks share a common submission interface and evaluation logic, additional datasets, modalities, and task types can be incorporated without changing the overall framework, allowing the benchmark to grow over time with new, clinically motivated tasks.

The variation across tasks further provided insights into where the UM-0 succeeded and for which task types the evaluation protocol was most restrictive. Within each vision domain, classification yielded the strongest baseline scores, with ISUP grading of prostate biopsies highest among all vision tasks (T1, 0.77) and lung nodule malignancy strongest in radiology (T2, 0.33), while dense prediction tasks clustered near the lower bound across both domains (T6–T8, T10–T11, all $\leq$ 0.04). This ordering aligns with the broader literature, in which frozen representations are most often evaluated, and perform most strongly, on classification, while few-shot dense prediction remains comparatively underexplored. This difference may in part reflect the constraints imposed by the evaluation protocol. To ensure that performance reflects representation transferability rather than task-specific tuning, UNICORN constrains all adaptors to lightweight architectures trained from scratch. For global tasks this is mild, as a small adaptor on a frozen embedding is a standard and effective probe. For dense tasks it is far more restrictive, removing the pretrained decoders that detection and segmentation conventionally rely on, such as the native encoder–decoder architectures of the segmentation variant of CT-FM (T10) and MRSegmentator (T6, T11). 

Although this adaptor constraint was applied uniformly across domains, its effect differed substantially, with the dense prediction tasks averaging 0.20 in pathology compared to 0.02 in radiology. This might be linked to the representations the adaptors operate on. The pathology encoder produces per-tile, two-dimensional features at a scale close to the targets, whereas the radiology encoders were not designed to expose comparable patch-level structure and operate on three-dimensional volumes, leaving the adaptor a harder spatial reconstruction from fewer aligned cues. This difficulty was compounded by the state of the field at the time of the challenge. With few general-purpose foundation models available for volumetric CT or MRI at the time of the challenge, the radiology baseline relied on models not originally intended as foundation encoders, such as a task-specific segmentation model for MR (MRSegmentator, T6, T11) and a natural-image encoder (DINOv2, T2), whereas mature encoder-style foundation models were already available in pathology. 

In addition to adaptor constraint and encoder fit, the task characteristics seem to influence the fit of the adaptor. The two pathology detection tasks diverged sharply under the same regime. Signet ring cell detection reached a moderate score (T5, 0.24), whereas mitotic figure detection failed almost entirely (T8, 0.00). Because both tasks used the same adaptor configuration, this gap points to task-specific factors, rather than difference in the evaluation protocol. The two tasks differ substantially in target-cell density. As shown in \autoref{fig:validation_distribution}, T5 contains relatively dense cellular targets, whereas T8 is characterized by sparse mitotic figures. A decoder optimized for dense objects performs poorly on sparse detection, potentially contributing to the divergence.

These results also show what a single baseline can and cannot establish about the causes of performance differences: the adaptor constraint, the encoder fit, and the task-specific characteristics all co-vary. For example, it remains unclear whether the dense-task gaps may close through adaptors better matched to each task or through encoders that yield more adaptable embeddings. Separating these effects requires evaluating a broader range of encoders and adaptors. To support this, public example datasets are released for each task, and the adaptor implementations and evaluation code are available in an open-source repository. Participants can evaluate their own encoders on the platform and contribute new adaptor strategies to it. With more submissions, these questions can be addressed directly.

Beyond the performance of UM-0, the framework itself represents a novel two-step evaluation platform. Supporting joint evaluation across such heterogeneous tasks required extending the Grand Challenge platform by handling multiple data modalities and prediction targets within a unified leaderboard, and enabling adaptor training within the Evaluation container. Whereas a standard challenge setup assumes a single homogeneous task, UNICORN introduces combined leaderboards that aggregate performance across tasks with distinct input types, output structures, and evaluation metrics. This design allows the evaluation of algorithms under a unified protocol despite the diversity of tasks. It supports task-specific leaderboards for debugging and detailed performance analysis, as well as combined leaderboards for assessing cross-task generalization. For combined leaderboards, the performance is summarized by the UNICORN Score.

The design of the UNICORN challenge was shaped by several practical constraints, particularly regarding data availability, computational resources, and architectural flexibility. Although the benchmark draws on datasets from multiple institutions, most test data originated from a single academic center, which limits assessment of how well submitted solutions generalize to other institutions, scanners, and patient populations. Computational constraints further shaped the evaluation protocol. Restrictions on submission count and per-case execution time excluded computationally expensive algorithms, but were necessary to keep large-scale evaluation feasible. Architectural flexibility was similarly limited. The separation of encoder and adaptor into distinct containers and the prohibition of pretrained adaptor weights prevented some state-of-the-art task-specific models from being submitted in their original form. These constraints keep the benchmark practically runnable but also narrow the range of modeling strategies that can currently be assessed.

UNICORN lays the groundwork for future efforts to evaluate generalist foundation models in medical imaging. By enabling standardized, few-shot evaluation across diverse tasks and modalities, it creates the conditions needed to study generalization under realistic clinical constraints.
Engagement from over 270 researchers across six continents reflects strong community interest in unified benchmarks and task-agnostic model development. As the benchmark evolves, new tasks, modalities, and centers can be integrated. Because many UNICORN tasks reuse datasets and metrics from earlier challenges, prior challenge-winning models can be re-evaluated within the same framework, allowing foundation models to be compared against strong task-specific baselines and clarifying when they match or complement specialized approaches. As medical foundation models continue to mature, UNICORN provides a stable and extensible protocol against which their progress toward genuine cross-domain generalization can be measured.

\section{Conclusion}
We introduce UNICORN, the first large-scale benchmark with a publicly available platform for two-step, few-shot evaluation of medical foundation models across diverse tasks, modalities, and clinical domains. With shared development datasets, and a multi-task infrastructure, it enables reproducible one-model-many-tasks comparison and combines task- and domain-specific performance with the UNICORN score. Moreover, it offers a detailed evaluation of each individual task and comparison of cross task- and cross domain performance. 

The UM-0 baseline demonstrates the end-to-end feasibility of the framework, achieving a UNICORN score of 0.38, which shows non-trivial performance across most tasks, while leaving room for future developments. 

The language tasks achieved the strongest performance, followed by global pathology vision tasks, whereas dense pathology vision tasks were more challenging. Radiology vision tasks presented additional challenges under the current encoder and adaptor setup.

By releasing the task descriptions, evaluation framework, UM-0 implementation, and evaluation code, UNICORN provides a common and evolving benchmark that supports model development and community collaboration. Such standardized evaluation is critical for developing robust, generalizable models suitable for clinical deployment. 
As medical foundation models continue to develop, UNICORN offers a stable basis for measuring progress toward robust cross-task and cross-domain generalization.

\printbibliography
\end{refsection}

\section{Declaration of generative AI and AI-assisted technologies in the manuscript preparation process}
During the preparation of this work, the authors used ChatGPT and Claude for clarity and consistency, and to support literature search. All citations were independently verified against the original sources before inclusion. After using this tool/service, the authors reviewed and edited the content as needed and take full responsibility for the content of the published article.

\section{Acknowledgments}
This research is supported by multiple public and private funding sources. The computational resources used for the UNICORN Challenge were partially supported by cloud credits provided by Amazon Web Services. Francesco Ciompi, Marina D'Amato, and Michelle Stegeman received funding from the Innovative Medicines Initiative 2 Joint Undertaking under grant agreement No 945358. This Joint Undertaking receives support from the European Union’s Horizon 2020 research and innovation program and EFPIA, www.imi.europe.eu. The views expressed are those of the authors only, and neither the IMI JU nor the Commission is liable for any use that may be made of the information contained therein.
Fennie van der Graaf was funded through a public-private project with funding from the Dutch Research Council (NWO), the Dutch Ministry of Economic Affairs, and MeVis Medical Solutions (Bremen, Germany). This study is part of the project ROBUST: Trustworthy AI-based Systems for Sustainable Growth with project number KICH3.L TP.20.006.
Luc Builtjes acknowledges support from RadboudAI.
Joeran S. Bosma received funding from HealthHolland (LSHM20103). The collaboration project is co-funded by PPP Allowance awarded by HealthHolland, Top Sector Life Sciences \& Health, to stimulate public-private partnerships. Views and opinions expressed are however those of the author(s) only and do not necessarily reflect those of the European Union or European Health and Digital Executive Agency (HADEA). Neither the European Union nor the granting authority can be held responsible for them. 
Rianne A. Weber received funding from the project 'OncoFuture: Hybrid AI for Efficient Cancer Diagnosis and Follow-Up Assessment' with file number NGF.1607.22.032 of the research programme AiNed Fellowship Grants which is (partly) financed by the Dutch Research Council (NWO).
For Francesco Ciompi, this publication is part of the project IGNITE with file number 18388 of the research programme Vidi - Applied and Engineering Sciences, which is (partly) financed by the Dutch Research Council (NWO). 
For Alessa Hering, this publication is part of the project OncoChange with file number 21121  of the research programme Veni - Applied and Engineering Sciences, which is (partly) financed by the Dutch Research Council (NWO).

\section{The UNICORN Consortium}
Michelle Stegeman$^{a,*}$,
Lena Philipp$^{b,*}$,
Fennie van der Graaf$^{b,*}$,
Marina D'Amato$^{a,*}$,
Cl\'ement Grisi$^{a,c,*}$,
Luc Builtjes$^{b,*}$,
Joeran S. Bosma$^{b,*}$,
Judith Lefkes$^{a,c,*}$,
Rianne A. Weber$^{b,*}$,
Alon Vigdorovits$^{f}$,
Anindo Saha$^{b}$,
Anna-Vibeke Laenkholm$^{g,h}$,
Bogdan Obreja$^{b}$,
Clara I. Sánchez Gutiérrez$^{d,e}$,
Chris van Run$^{b}$,
Christian Roest$^{i}$,
David Tellez$^{a}$,
Denis Larsimont$^{j}$,
Dennis Rouw$^{k}$,
Derya Yakar$^{l,i}$,
Dieter J.E. Peeters$^{m,n}$,
Dr\'e Peeters$^{b}$,
Elisabeth Ida Specht Stovgaard$^{o,p}$,
Elise M.G. Taken$^{b}$,
Ernst Th. Scholten$^{b}$,
Gijs Stultiens$^{b}$,
Harm van Zeeland$^{a}$,
Harry Haynes$^{q}$,
Inge M. E. Smit$^{b}$,
Iris D. Nagtegaal$^{a}$,
Ivan R. Slootweg$^{a}$,
Jasper J. Twilt$^{b}$,
Jasper van der Graaf$^{b}$,
Jeroen Geerdink$^{r}$,
Jeroen Veltman$^{s}$,
Job van Susante$^{t}$,
Joep Bogaerts$^{a}$,
Joey M. A. Spronck$^{a}$,
Jos J. Immerzeel$^{u}$,
Jurgen J. Fütterer$^{b}$,
Karlijn Rutten$^{b}$,
Khrystyna Faryna$^{a}$,
Laura Comerma Blesa$^{v}$,
Leander van Eekelen$^{a}$,
Lee Cooper$^{w}$,
Leslie Tessier$^{a,x}$,
Ligia Craciun$^{j}$,
Mar Navarro-Padilla$^{b}$,
Mart van Rijthoven$^{a}$,
Maschenka Balkenhol$^{a,y}$,
Matthieu J.C.M. Rutten$^{b,z}$,
Mattijs Elschot$^{aa,ab}$,
Max J. J. de Grauw$^{b}$,
Megan Schuurmans$^{b}$,
Michiel Simons$^{a}$,
Miriam Groeneveld$^{b}$,
Mohamed Amgad$^{w}$,
Mohammed R. S. Sunoqrot$^{aa,ab}$,
Nat\'alia Alves$^{b}$,
Nefise Uysal$^{a}$,
Nikolas Lessmann$^{b}$,
Noa Antonissen$^{b}$,
Paulo Guilherme de Oliveira Salles$^{ac}$,
Paul K. Gerke$^{b}$,
Quintin van Lohuizen$^{i}$,
Rachel S. van der Post$^{a}$,
Robert Jan Kroeze$^{ad}$,
Robin Lomans$^{a}$,
Shoko Vos$^{a}$,
Stefan J. Fransen$^{i}$,
Tone F. Bathen$^{aa,ab}$,
Valentina Angerilli$^{a}$,
Ward Hendrix$^{b}$,
Witali Aswolinskiy$^{a}$,
Zaigham Saghir$^{af}$,
James A. Meakin$^{b}$,
Anne Mickan$^{b}$,
Thomas Koopman$^{d,e}$,
Mathias Prokop$^{b}$,
Ewoud J. Smit$^{b}$,
Geert Litjens$^{a,c}$,
Jeroen van der Laak$^{a}$,
Bram van Ginneken$^{b}$,
Maarten de Rooij$^{b}$,
Henkjan Huisman$^{b}$,
Colin Jacobs$^{b}$,
Francesco Ciompi$^{a,\dagger}$,
Alessa Hering$^{b,\dagger}$

\vspace{1em}

\noindent \textbf{Affiliations:}\\
$^{a}$ Department of Pathology, Radboud University Medical Center, Nijmegen, The Netherlands \\
$^{b}$ Department of Medical Imaging, Radboud University Medical Center, Nijmegen, The Netherlands \\
$^{c}$ Oncode Institute, Utrecht, the Netherlands \\
$^{d}$ Informatics Institute, Faculty of Science, University of Amsterdam, Amsterdam, The Netherlands \\
$^{e}$ Departments of Biomedical Engineering and Physics, University Medical Center Amsterdam, Amsterdam, The Netherlands \\
$^{f}$ Faculty of Medicine, University of Oradea in Oradea, Romania \\
$^{g}$ Department of Pathology, Herlev-Gentofte Hospital, Herlev, Denmark \\
$^{h}$ Department of Clinical Medicine, Copenhagen University, Copenhagen, Denmark
$^{i}$ Department of Radiology, University Medical Center Groningen, Groningen, The Netherlands \\
$^{j}$ Department of Pathology, Institut Jules Bordet, Université Libre de Bruxelles (ULB), Brussels, Belgium \\
$^{k}$ Department of Radiology and Nuclear Medicine, Martini Hospital, Groningen, The Netherlands \\
$^{l}$ University of Groningen, Groningen, Netherlands \\
$^{m}$ Department of Pathology, Antwerp University Hospital, Edegem, Belgium \\
$^{n}$ Department of Pathology, AZ Sint-Maarten, Mechelen, Belgium \\
$^{o}$ Department of Pathology, Herlev and Gentofte Hospital, Herlev, Denmark \\
$^{p}$ Department of Clinical Medicine, University of Copenhagen, Denmark \\
$^{q}$ Great Western Hospitals NHS Foundation Trust, Swindon, United Kingdom \\
$^{r}$ Department of Health \& Information Technology, Ziekenhuisgroep Twente, Almelo, The Netherlands \\
$^{s}$ Department of Radiology, Ziekenhuisgroep Twente, Almelo, The Netherlands \\
$^{t}$ Department of Radiology and Nuclear Medicine, Rijnstate Hospital, Arnhem, The Netherlands \\
$^{u}$ Andros Clinics, The Netherlands \\
$^{v}$ Department of Pathology, Hospital del Mar, Barcelona, Spain \\
$^{w}$ Department of Pathology, Northwestern University Feinberg School of Medicine, Chicago, IL, United States of America \\
$^{x}$ Institut du cancer de l'Ouest, University hospital, Angers France \\
$^{y}$ Department of Pathology, Canisius Wilhelmina Hospital, Nijmegen, the Netherlands \\
$^{z}$ Department of Radiology, Jeroen Bosch Hospital, ’s-Hertogenbosch, The Netherlands \\
$^{aa}$ Department of Radiology and Nuclear Medicine, St Olavs Hospital, Trondheim University Hospital, Trondheim, Norway \\
$^{ab}$ Department of Circulation and Medical Imaging, The Norwegian University of Science and Technology -- NTNU, Trondheim, Norway \\
$^{ac}$ Anatomical Pathology Service, Instituto Mário Penna, Belo Horizonte, Brazil \\
$^{ad}$ Department of Radiology, Sint Maartenskliniek, Nijmegen, The Netherlands \\
$^{af}$ Department of Medicine, Section of Pulmonary Medicine, Herlev-Gentofte Hospital, Hellerup, Denmark \\

\section{authorship contribution statement}
Michelle Stegeman: Writing - Original Draft, Methodology, Software, Data Curation. Lena Philipp: Writing - Original Draft, Methodology, Software, Data Curation. Fennie van der Graaf: Writing - Original Draft, Methodology, Software, Data Curation. Marina D’Amato: Writing - Review \& Editing, Methodology, Software, Data curation. Clément Grisi: Writing - Review \& Editing, Methodology, Software, Data curation. Luc Builtjes: Writing - Review \& Editing, Methodology, Software, Data curation. Joeran S. Bosma: Writing - Original Draft, Methodology, Software, Data curation. Judith Lefkes: Writing - Review \& Editing, Methodology, Software, Data curation. Rianne A. Weber: Writing - Review \& Editing, Methodology, Data curation
James A. Meakin: Writing - Review \& Editing, Software. Thomas Koopman: Software. Anne Mickan: Software. Mathias Prokop: Supervision. Ewoud J. Smit: Supervision. Geert Litjens: Supervision. Jeroen van der Laak: Supervision. Bram van Ginneken: Writing - Review \& Editing, Supervision. Maarten de Rooij: Supervision. Henkjan Huisman: Supervision. Colin Jacobs: Writing - Review \& Editing, Supervision. Frédérique Meeuwsen: Supervision. Francesco Ciompi: Writing - Review \& Editing, Conceptualization, Methodology, Data curation, Supervision, Project administration. Alessa Hering: Writing - Original Draft, Methodology, Data curation, Supervision, Project administration.

\newpage
\begin{refsection}
% Reset counters
\setcounter{section}{0}
\setcounter{figure}{0}
\setcounter{table}{0}
\setcounter{page}{1}

\renewcommand\thesection{Supplemental \arabic{section}}
\renewcommand\theHsection{supp.\arabic{section}}

\renewcommand\thetable{A.\arabic{table}}
\captionsetup[table]{name=Table}
\crefname{table}{Table}{Tables}
\Crefname{table}{Table}{Tables}

\renewcommand\thefigure{A.\arabic{figure}}
\captionsetup[figure]{name=Figure}  
\crefname{figure}{Figure}{Figures} 
\Crefname{figure}{Figure}{Figures} 

\renewcommand\theequation{A.\arabic{equation}}

\section{Tasks}
\label{app:tasks}
\subsection{Vision -- classification}
\subsubsection*{T1: Classifying H\&E-stained prostate biopsies into ISUP scores}
This task involves classifying prostate biopsies as either benign or into one of the five International Society of Urological Pathology (ISUP) grade groups. The dataset includes a total of 308 H\&E-stained whole-slide images (WSI) from prostate cancer patients in screening programs and hospital evaluations. Data was collected from two cohorts: 195 whole-slide images from Radboudumc (2012--2017) for the validation set and for the test set, 113 whole-slide images from six hospitals, the Rennes University Hospital (France), Radboudumc (Netherlands), Institute of Pathology and Molecular Immunology of the Univeristy of Porto (Portugal), Memorial Hospitals Istanbul (Turkey), Foch Hospital (France), and Isala Zwolle (Netherlands), collected as part of the Evaluating AI-based Gleason grading in-the-wild study \cite{gginthewild} to ensure a diverse and representative set of prostate biopsy slides from routine clinical practice.

The first cohort is sampled from the test set of the PANDA challenge \cite{bulten2022artificial}\footnote{\url{https://panda.grand-challenge.org/}}. The reference standard was established through a three-round grading procedure involving three expert uropathologists, following the ISUP 2014 guidelines. In the first round, all cases were graded independently. If at least two pathologists agreed on the ISUP grade group, the majority vote was accepted, including cases where the disagreement was limited to the Gleason pattern order (e.g., 4+5 vs. 5+4) or minor differences in score interpretation. Cases with a disagreement on malignancy were always flagged for further review. In the second round, cases without consensus were regraded by the outlier pathologist, i.e., the one whose score deviated most from the other two. During this step, the anonymous scores from the other two pathologists were made available to aid re-evaluation. In the third round, any remaining cases without agreement were reviewed in a consensus meeting. For the second cohort, a single representative biopsy per case was selected by a pathology resident under the supervision of a uropathologist. A reader study involving 10 pathologists established the reference standard. For each case, the final label was assigned through a two-step process: first, the sample was categorized as benign or cancerous based on the 10 grades. If deemed cancerous, a majority vote among the cancer grades determined the final label. In case of ties, the highest Gleason grade group was assigned.

Data was obtained in various formats, but is provided in the benchmark as multi-resolution .tif files with a starting resolution of 0.50 microns per pixel, along with a binary tissue mask (0=background, 1=tissue). For each case, the expected prediction is a single integer ranging from 0 to 5. Task performance is assessed using Cohen's quadratic weighted kappa.

\subsubsection*{T2: Classifying lung nodule malignancy in CT}
The main goal of this task is to predict a nodule malignancy probability for a nodule candidate, indicating either a low or high risk of malignancy. The test data is from the Danish Lung Cancer Screening Trial (DLCST)\cite{pedersen2009dlst}, which includes low-dose CT thorax-abdomen scans from between 2004 and 2010, containing 533 nodules from 533 participants. Participants aged 50--70 years at the time of inclusion and with a minimum of 20 pack-years of smoking were included. Each case contains one nodule block of $128 \times 128 \times 64$ voxels from the low-dose CT thorax-abdomen scans provided in the MHA file format. No additional preprocessing was performed. The first occurrence of a malignant nodule for each cancer was selected, and then one malignant nodule per participant was randomly selected and annotated. A positive cancer case refers to a scan with at least one malignant nodule, confirmed by histopathological examination. Benign nodules are only provided from non-lung cancer scans with at least 2 years of imaging follow-up. The nodule locations were annotated as center coordinates by two experienced chest radiologists. The scans were acquired with a 16-slice Philips Mx 8000, and scans were taken with the patient supine after full inspiration using a low-dose technique (120 kV and 40 mA). 

For each case, the prediction is expected to be a continuous model probability score between 0 and 1. The task will be assessed using the area under the receiver operating characteristic curve (AUROC) to evaluate performance in distinguishing between benign and malignant nodules. 

This dataset is similar to a subset of the testing dataset used in the LUNA25 challenge \cite{peeters_luna25_2025}\footnote{\url{https://luna25.grand-challenge.org/}}, with key differences being that the UNICORN task 2 testing set contains nodules of all sizes, not only of the indeterminate size range, and that it is not cancer-enriched. 

\subsubsection*{T4: Predicting slide-level tumor proportion score in NSCLC IHC-stained WSI}
The goal in this task is to predict the tumor proportion score (TPS) in PD-L1 immunohistochemistry-stained whole-slide images from non-small cell lung cancer (NSCLC). The dataset includes a total of 639 WSIs from 505 lung cancer patients from Radboudumc, including biopsies from both primary lung tumors and metastatic sites. 

Each individual case is provided on the platform as a .tif file with a resolution of 0.50 microns per pixel, accompanied by a binary tissue mask (0=background, 1=tissue). Positive control tissue was excluded from the WSIs, as its sole purpose is to verify the success of the immunohistochemical staining procedure. The reference standard for each case is a binned TPS value, extracted from pathology reports and based on visual estimates by experienced lung pathologists. In clinical settings, these estimates are typically categorized into three bins: less than one percent, one percent to less than fifty percent, and fifty percent or greater.
The expected prediction for each case is a single integer (0,1,2) reflecting one of the 3 bins. Performance is evaluated using Cohen’s quadratic weighted kappa between predicted and reference TPS bins.

\subsection{Vision -- regression}
\subsubsection*{T3: Predicting the time to biochemical recurrence in H\&E-stained prostatectomies}
This task aims to estimate the risk of biochemical recurrence in prostate cancer patients after a radical prostatectomy. Biochemical recurrence is defined as PSA $> 0.2$ ng/mL on two or more occasions after having previously reached an undetectable level post-prostatectomy. The dataset includes H\&E-stained whole prostatectomy slides for 570 patients, with follow-up for prostate-specific antigen (PSA) blood measurements. Data was collected from two cohorts: 149 patients from Radboudumc (Netherlands), and 421 patients from Instituto Mário Penna (Brazil).

Each case is provided as multi-resolution .tif files, with a starting resolution of 0.50 microns per pixel along with a binary tissue mask (0=background, 1=tissue). The label for each case indicates whether biochemical recurrence occurred (0=no biochemical recurrence, 1=biochemical recurrence), and the time in years to recurrence or the last follow-up. No human annotators were involved, as the biochemical recurrence is used as the primary endpoint. 

For each case the prediction should be a continuous time estimate (float) for the time to biochemical recurrence (in years). Performance is evaluated using the censored concordance index.

Task 3 is adapted from the LEOPARD challenge \cite{leopard}\footnote{\url{https://leopard.grand-challenge.org/}}. Unlike the LEOPARD challenge, which relied on local training with large datasets, this task follows a few-shot learning paradigm, with training performed on-platform.

\subsection{Vision -- detection}
\subsubsection*{T5: Cell detection of signet ring cells in H\&E-stained WSI of gastric cancer}
This task aims to detect signet ring cells in H\&E-stained whole-slide images of gastric cancer. The dataset includes 478 regions of interest (ROIs) extracted from 259 WSIs of 35 patients from Radboudumc, all scanned using the same scanner. Each ROI is considered one case and is saved as an individual .tif file at a starting resolution of 0.5 microns per pixel. 

ROIs with diffuse gastric cancer tumor lesions, including pre-invasive lesions, were selected by a resident pathologist. All signet ring cells were annotated as (x,y) coordinates by student annotators, with all annotations subsequently verified by a pathologist with expertise in gastric cancer.

The expected prediction for each case consists of predicted (x,y) coordinates of detected signet ring cells within the case. Performance is evaluated using the F1 score. If multiple predicted points hit one ground truth point, this is counted as 1 true positive and 0 false negatives. If one predicted point is a hit for N ground truth points, this is counted as 1 true positive and N-1 false negatives.

\subsubsection*{T6: Detecting clinically significant cancer in prostate MRI exams}
The objective in this task is to perform 3D detection of clinically significant prostate cancer lesions with an ISUP score $\geq$ 2, using biparametric MRI scans.
The patient population includes individuals suspected of harboring clinically significant prostate cancer based on elevated PSA levels or abnormal DRE findings, with no prior treatment or history of ISUP $\geq$ 2 findings.
Each case within the benchmark comprises axial T2-weighted (T2W) scans, axial high b-value ($\geq 1000 s/mm^2$) diffusion-weighted imaging (DWI), and axial apparent diffusion coefficient (ADC) maps.  
Test data consists of 400 cases and includes all three imaging sequences (axial T2W, DWI, and ADC). Data was collected from Radboudumc, Ziekenhuisgroep Twente, Prostaat Centrum Noord-Nederland, and St. Olav’s Hospital in Norway, using Siemens or Philips scanners.  
The reference standard for validation and testing incorporates histopathological evidence (ISUP $\geq$ 2 for positives, ISUP $\leq$ 1 or MRI PI-RADS $\leq$ 2 with $\geq$ 3 years follow-up for negatives). Annotations for cases were performed by trained investigators under expert supervision. These annotations were based on MRI exams, diagnostic reports, and pathology specimens, undergoing quality control and independent cross-checking at the coordinating medical center.  
The evaluation metric is the average of the area under the receiver operating characteristic curve (AUROC) and the area under the precision-recall curve (AP).
Task 6 is adapted from the PI-CAI challenge \cite{saha2024artificial}\footnote{\url{https://pi-cai.grand-challenge.org/}}. Validation and test datasets mirror those used in PI-CAI, with the test dataset mirroring the reader study subset, but omitting clinical parameters and coronal/sagittal T2 images.

\subsubsection*{T7: Detecting lung nodules in thoracic CT}
The main goal of this task is to accurately detect pulmonary nodules in clinical routine chest CT scans. 
The test cohort of 83 patients includes clinical scans collected in the period of 2018-2020 at Radboudumc. The test set is sourced from one of the hospitals in a previous study \cite{hendrix2023deep}, however, from 100 patients the test set was filtered to 83. We filtered scans for which the majority (3/5) of the radiologists agreed on the presence of the nodule, to create a test set with a reliable reference standard. Patients include those receiving thorax CTs as part of clinical care, who are above 18 years old, and in whom a nodule was incidentally detected. Both contrast-enhanced and non-contrast CT scans are included. Data acquisition involved scanners from various manufacturers, including Philips, Siemens, and Canon models.  
The dataset was annotated by a panel of five experienced thoracic radiologists using specialized in-house software to identify and measure intrapulmonary nodules and generate coordinates. Each case contains one CT volume provided in .mha format (512 x 512 px), with no pre-processing applied. The model predictions are evaluated using sensitivity and false positive rate per scan. To be a true positive, the lesion candidate should be located within a distance $R$ of the nodule center, where $R$ is set to the equivalent diameter of the nodule divided by 2. Lesion candidates that do not correspond to any nodule are considered false positives. Lesion candidates corresponding to irrelevant findings are ignored during the evaluation and are not considered either false positives or true positives. Performance is assessed in the same way as the LUNA16 challenge, using a Free Receiver Operating Characteristic (FROC) analysis and the competition performance metric (CPM) is defined as the average sensitivity at 7 predefined false positive rates: 1/8, 1/4, 1/2, 1, 2, 4, and 8 false positives per scan. If none of the candidates result in a true positive, a score of 0 for the CPM will be assigned. Task 7 is inspired by the LUNA16 challenge \cite{SETIO2017}\footnote{\url{https://luna16.grand-challenge.org/}}.

\subsubsection*{T8: Cell detection of mitotic figures in breast cancer H\&E-stained WSIs}
The goal of this task is to accurately detect mitotic figures in H\&E-stained whole-slide images of triple-negative breast cancer. The dataset includes 400 regions of interest (ROIs) extracted from 11 whole-slide images from patients from Radboudumc. Each ROI is considered one case and extracted and saved as an individual .tif file at a resolution of 0.25 microns per pixel. All ROIs are selected using random probability sampling, where ROIs with higher mitotic count are more likely to be sampled. Annotations of all mitotic figures are provided as the (x, y) coordinates within the case. Each mitotic figure has been validated using a PHH3 marker, clearly highlighting the mitotic cells, on an immunohistochemistry-stained image. The clearly marked cells by the PHH3 marker were used for registration to the corresponding H\&E-stained image.

The expected prediction for each case should be all the predicted (x,y) coordinates of the mitotic figures. The performance is evaluated using the F1-score. If multiple predicted points hit one ground truth point, this is counted as 1 true positive and 0 false negatives. If one predicted point is a hit for N ground truth points, this is counted as 1 true positive and N-1 false negatives. 

\subsection{Vision -- segmentation}
\subsubsection*{T9: Segmenting ROIs in breast cancer H\&E-stained WSIs}
Within this task, the objective is to segment breast tissue into tumor, stroma, and other in H\&E-stained whole-slide images. The dataset includes 33 regions of interest (ROIs) from 11 whole-slide images from Radboudumc and the Jules Bordet Institute. Each case is an ROI of approximately 500x500 microns extracted and saved as individual .tif files at a resolution of 0.50 microns per pixel. Each ROI was annotated by five board-certified breast pathologists from the tumor-infiltrating lymphocyte working group. All data was adopted from the TIGER challenge \cite{tiger}\footnote{\url{https://tiger.grand-challenge.org/}}, however, only a subset of sequestered data was used in the form of ROIs instead of WSIs.

The prediction for each case should be a 2D array with pixel-level segmentation of four classes (0=background, 1=tumor, 2=stroma, 3=other). Performance is assessed using the Dice coefficient. 

\subsubsection*{T10: Segmenting lesions within ROIs in CT}
This task focuses on segmenting lesions within ROIs in chest-abdomen-pelvis CT examinations. The goal of this task is to segment 3D masks of lesions in CT scans.
The dataset includes CT sub-images of lesions from 725 cases (46.5\% female) sourced from Radboudumc and Jeroen Bosch Ziekenhuis in the Netherlands. 
Each case comprises a volume of interest (VOIs) of $256 \times 256 \times 128$ voxels. Each VOI centers on a lesion voxel to simulate a radiologist’s click. 
Where necessary, volumes were padded with the minimum intensity value of the VOI minus one. Each VOI contains a single annotated lesion. 
Three annotators independently annotated each lesion in 3D, and the final mask was determined by majority vote. 
This mask was reviewed and corrected where necessary by an experienced radiologist. 
Segmentation performance (SP) is measured using the Sørensen-Dice coefficient. Additional metrics include symmetric mean absolute percentage error for long- and short-axis measurements (LAE and SAE).
The final composite score is computed as: 
    \[
    \text{CS} = 0.888 \cdot \text{SP} + 0.056 \cdot \text{LAE} + 0.056 \cdot \text{SAE}
    \]
This test set was previously used in the ULS challenge \cite{uls}\footnote{\url{https://uls23.grand-challenge.org/}}. The evaluation simplifies the original challenge evaluation by omitting the robustness score.

\subsubsection*{T11: Segmenting three anatomical structures in lumbar spine MRI}
The goal of this task is to segment vertebrae, intervertebral discs, and the spinal canal using lumbar MRI scans.
The data includes MRI studies of patients with a history of low back pain, retrospectively collected from four Dutch hospitals in the Netherlands: one university medical center, two regional hospitals, and one orthopedic hospital.
The data was acquired between January 2019 and March 2022 and includes 97 MRI series from 39 patients (62\% female).  
Each case is one patient examination that consists of up to three sagittal MRI series, which are either T1-weighted or T2-weighted (regular resolution, or high resolution generated using a SPACE sequence). 
The voxel size of these sequences ranges from $0.90 \times 0.47 \times 0.47$~mm to $9.63 \times 1.06 \times 1.23$~mm.
All visible vertebrae (excluding the sacrum), intervertebral discs, and the spinal canal were manually annotated.  
The annotations were performed by a medical trainee who was trained and supervised by both a medical imaging expert and an experienced musculoskeletal radiologist.  
The segmentation performance is evaluated using the Dice coefficient measured in 3D. Dice scores are computed for each individual anatomical structure (vertebrae, intervertebral discs, and the spinal canal). Instance-wise scores are then averaged to obtain an overall Dice score.

This test set was previously used in the SPIDER challenge \cite{van2024lumbar}\footnote{\url{https://spider.grand-challenge.org/}}.

\subsection{Language -- classification}
The language tasks (tasks 12 to 19) are derived from tasks in the DRAGON challenge \cite{bosma2025dragon}. For some tasks, the evaluation data is exactly the same as in the DRAGON challenge, while for others, adaptations have been made to align better with generative LLMs, as detailed below. For each task, 32-48 few-shot examples are provided to facilitate in-context learning. All reports were anonymized using the Hide-in-plain-sight (HIPS) \cite{carrell2013hiding} algorithm to remove sensitive patient information. Apart from anonymization, no data preprocessing was performed for any of these tasks.

\subsubsection*{T12: Predicting histopathology sample origin}
The objective in this task is to predict from which organ the material was taken (lung, lymph node, bronchus, liver, brain, bone, or other), as described in the Dutch pathology report. The challenge cohort includes patients suspected of having non-small cell lung cancer between January 1, 2016, and December 31, 2022. The same test and validation sets were used as in Task06 from the DRAGON challenge \cite{bosma2025dragon}. For adaptation, 48 few-shot examples were sampled from the training data, without overlap between the validation and test phases, resulting in a total of 608 Dutch pathology reports being included, all originating from Radboudumc. The labels were assigned by trained student assistants based on manual review of the reports. The labels for this task are categorical and the frequency of the labels is unbalanced. The labels are not ordinal. Therefore, model performance is assessed using the unweighted Cohen's Kappa.

\subsubsection*{T13: Classifying pulmonary nodule presence}
The objective in this task is to determine whether a pulmonary nodule is described in the radiology report or not. Chest CT reports from two Dutch hospitals, acquired between 1 January 2008 and 31 December 2019 from Radboudumc and Jeroen Bosch Ziekenhuis were used \cite{t13_test_hendrix}. The same test and validation sets were used as in Task07 from the DRAGON challenge \cite{bosma2025dragon}. For adaptation, 48 few-shot examples were sampled from the training data, without overlap between the validation and test phases, resulting in a total of 596 Dutch reports being used from 596 unique patients. The dataset reflects real-world class distributions. The test reports were annotated by a radiologist with 26 years of experience. Each report received a binary label indicating the presence or absence of pulmonary nodules. The area under the receiver operating characteristic curve is used to evaluate model performance.

\subsubsection*{T14: Classifying kidney abnormality}
The objective in this task is to identify the presence of significant kidney abnormalities based on the radiology report. The same test and validation sets were used as in Task03 from the DRAGON challenge \cite{bosma2025dragon}. For adaptation, 48 few-shot examples were sampled from the training data, without overlap between the validation and test phases, resulting in a total of 404 Dutch radiology reports collected from patients who underwent abdominal CT scans at Radboudumc. Reports were manually analyzed and labeled as either ``normal'' or ``abnormal'' by a native Dutch-speaking PhD candidate. Abnormalities include renal cell carcinoma, angiomyolipoma, cysts, kidney stones, conjoined kidneys, cases with partial or full nephrectomy, and several other rare abnormalities. Evaluation is performed using the area under the receiver operating characteristic curve.

\subsubsection*{T15: Predicting hip Kellgren-Lawrence score}
The objective in this task is to classify the radiology report by the Hip Kellgren-Lawrence score for hip osteoarthritis per hip joint, ranging from 0 to 4 with 0 (no osteoarthritis), 1 (doubtful osteoarthritis), 2 (minimal osteoarthritis), 3 (moderate osteoarthritis), and 4 (severe osteoarthritis), or indicate the hip has a prosthesis, or the scoring is not applicable to that hip. The test set is a subset of 108 cases from the original test set from Task018 in the DRAGON challenge \cite{bosma2025dragon}, the validation set is a subset of 100 cases from the original validation set, and for adaptation, 32 few-shot examples were sampled from the original testing data, without overlap between the validation and test phases resulting in a total of 272 Dutch radiology reports for hip osteoarthritis collected at Radboudumc. Reports were manually annotated by a trained investigator, with difficult cases being reviewed by a radiologist. The labels in the dataset are imbalanced, and while the Kellgren-Lawrence score is ordinal, the additional categories are not. Performance is assessed by pooling the predictions for the left and right hip, and calculating the unweighted Cohen's Kappa for the combined predictions. 

\subsubsection*{T16: Classifying colon histopathology diagnosis}
The objective in this task is to predict whether the specimen was obtained from 1) biopsy or polypectomy, and whether the pathologist rated the specimen as 2) cancer, 3) high-grade dysplasia (hgd), 4) hyperplastic polyps, 5) low-grade dysplasia (lgd), 6) non-informative (ni), or 7) serrated polyps. Each of these seven properties is binary, and multiple can be present per block. Each report conclusion was annotated by native Dutch data analysts under the supervision of two gastrointestinal pathologists. The test set is a subset of 500 cases from the original test set in Task015 from the DRAGON challenge \cite{bosma2025dragon} with sampling strategy to ensure at least 20 positive cases per label, the validation set is a subset of 250 cases from the original validation set with the same sampling strategy, and for adaptation, 48 few-shot examples were sampled from the original training data, without overlap between the validation and test phases and with at least three positive cases per label, resulting in 846 Dutch pathology reports from patients diagnosed with histopathological conditions of the colon, from biopsies performed between January 1, 2000, and December 31, 2009, at Radboudumc. For patients with multiple visits during this time frame, only the first visit was included. The model's output is expected as a JSON file, containing per case the labels for all eight characteristics. Model performance is evaluated by calculating the area under the receiver operating characteristic curve (AUROC) for each class separately. The overall performance of the model is defined as the average of the AUROC values across all classes.

\subsection{Language -- regression}
\subsubsection*{T17: Predicting lesion size measurements}
The objective in this task is to extract the size of three types of lesions which are presented in the report. The lesions to be extracted were offered in the preamble of each report. Data is sourced from three tasks of the DRAGON challenge \cite{bosma2025dragon}: pulmonary nodules, RECIST target lesions, and pancreatic ductal adenocarcinomas (PDAC), representing real-world cases of diagnostic procedures. This is a combined subset of Task022, Task023 and Task024 of the DRAGON challenge \cite{bosma2025dragon} cohort, using a total of 636 reports. 
The sequestered test set includes 298 cases distributed across these cohorts: 32 cases to extract the pulmonary nodule size, 119 cases to extract RECIST lesion sizes, and 147 cases to extract the PDAC size. Annotations were manually created by trained investigators or radiologists, ensuring accurate lesion size measurements. Each dataset has specific prediction objectives, such as determining the largest diameter, short axis, or longest axis of lesions, depending on the type of lesion described in the radiology report. Both the report text and the annotations are provided for the few shot cases in a JSON file.  
Lesion size ranges vary by cohort, with pulmonary nodules measuring between 1--30 mm, RECIST lesions between 4--166 mm, and PDAC diameters between 6--130 mm, as determined by the development data (values in the test set are sequestered), with most values falling within specific diagnostic ranges. Evaluation uses the Robust Symmetric Mean Absolute Percentage Error Score (RSMAPES) with a 4 mm tolerance to ensure diagnostically relevant evaluation.

\subsubsection*{T18: Predicting prostate volume, PSA, and PSA density}
The objective in this task is to predict the prostate volume, prostate specific antigen (PSA) level, and the PSA density, based on the radiology report. This task was derived from Task019, Task020, and Task021 the DRAGON challenge \cite{bosma2025dragon} cohort. Dutch prostate MRI reports were collected from University Medical Center Groningen, Antoni van Leeuwenhoek Ziekenhuis, and Radboudumc. The combined dataset includes 846 cases. Labels were manually annotated by trained investigators and consist of three continuous variables. The prostate volume ranges from 4.0 to 470 cm³, with 95\% of values between 24 and 181 cm³, as determined based on the original task's development data. The PSA level ranges from 0 to 870 ng/mL, with 95\% of values between 2 and 35 ng/mL, as determined based on the original task's development data. The PSA density ranges from 0 to 171 ng/mL², with 95\% of values between 0.03 and 0.6 ng/mL², as determined based on the original task's development data. Performance for this task is measured using the average of the RSMAPES metric \cite{bosma2025dragon} per clinical variable, with epsilon thresholds set at 4 cm³ for prostate volume, 0.4 ng/mL for PSA level, and 0.04 ng/mL² for PSA density.
Task 18 is derived from Task019 (prostate volume extraction), Task020 (prostate specific antigen extraction), and Task021 (prostate specific antigen density extraction) of the DRAGON Challenge \cite{bosma2025dragon}.

\subsection{Language -- named entity recognition}
\subsubsection*{T19: Anonymizing report}
The objective in this task is to identify and tag personally identifiable information (PII) within reports, including dates, personal identifiers, report identifiers, locations, clinical trial names, times, and ages. 
Pathology and radiology reports were sourced from two hospitals (Radboudumc, Antoni van Leeuwenhoek Ziekenhuis) as part of Task025 of the DRAGON challenge \cite{bosma2025dragon}. The test set is a subset of 400 cases from the original test set in the DRAGON challenge \cite{bosma2025dragon}, the validation set is a subset of 200 cases from the original validation set, and for adaptation, 48 few-shot examples were sampled from the original training data, without overlap between the validation and test phases, resulting in a total of 696 cases being included.
The reports were annotated semi-automatically by processing the reports with a tool developed in-house (i.e., a rule-based system) and were manually verified and corrected by one of three trained investigators. 
Performance is assessed using the blended redaction F1 score. This metric combines a strict character-level F1 score checking the correct tag type with a binary character-level F1 score determining whether the span was redacted at all. These values are weighted 0.7 and 0.3 in the final score, respectively.

\subsection{Vision-language -- caption generation}
\subsubsection*{T20: Generating caption from WSI}
This task involves generating a caption for each case that encapsulates the relevant clinical information, such as tissue type and diagnosis, from the whole-slide image (WSI) of colon and cervix biopsies and polyps. The dataset consists of 392 WSI-report pairs from 361 unique patients derived from the ExaMode project, where each case is a WSI linked to its corresponding Dutch pathology reports. The ground truth consists of the conclusion section from the original Dutch pathology report associated with each WSI, which captures the pathologist’s final diagnostic interpretation.

The dataset includes 202 colon cases and 190 cervix cases. To ensure diagnostic diversity and mitigate class imbalance, cases were sampled to evenly represent clinically relevant categories based on standardized protocols for cervix and colon biopsies. For cases with multiple co-occurring diagnoses (e.g., CIN3 with carcinoma), the most clinically severe label guided inclusion.

Each case is provided in the benchmark as .tif WSI, along with its corresponding tissue mask and task description. The algorithms are expected to analyze the WSI, identify key features such as tissue type, cellular morphology, and pathological findings, and produce a clear, clinically meaningful Dutch textual conclusion per case. Performance is evaluated as the average of five metrics: BLEU-4, ROUGE-L, CIDEr, METEOR, and BERTScore.

\section{Challenge Phases}
\label{app:challenge_phases}
To support comprehensive benchmarking, the UNICORN Challenge was arranged in sequential phases, each with specific goals and evaluation criteria.

\subsection{Off-Platform Development Phase}
Given the limited number of permitted on-platform submissions, strong local development was essential for identifying promising approaches before on-platform evaluation. In the off-platform development phase, users could freely develop and validate their models and aggregation strategies outside the Grand Challenge platform. To support this, we publicly released example datasets for each task, typically sourced from public data and reformatted to match the format and structure of the sequestered data in the on-platform check, validation, and test phases \cite{damato_unicorn_2025}. The examples supported local solution development while ensuring compatibility with the platform's inputs and outputs. References to existing public datasets were also provided to facilitate further local experimentation, as provided in \Cref{sup_tab:dataset_info}.

\subsection{Check phase} 
The check phase was designed to verify that submission pipelines execute correctly on the platform, identify potential issues, and help teams become familiar with the platform’s submission process. Submissions in this phase did not contribute towards the final leaderboard, as the results were not always reflective of model performance because only a handful of cases were used. For this phase, a limited but carefully curated set of few-shot and evaluation cases was selected, including potential edge cases such as extremely large or unusually small images, which are most likely to trigger errors or timeouts. Passing the check phase was a prerequisite for submitting to the corresponding validation or test phase. This requirement contributed to only permitting robust and well-integrated pipelines to progress to later stages, reducing the risk of runtime failures during validation and test evaluations. In addition, users had access to the log file on the platform and could submit an unlimited number of times to the Check Phase, hence conserving their submissions on the validation phase for algorithm improvements rather than debugging. By being able to debug their errors early in the challenge, we were able to save compute costs, as opposed to users discovering errors only in the validation and test phase, which have large numbers of cases. 

\subsection{Validation phase}
\label{app:challenge_phase_validation}
The validation phase was designed to help users assess their model and compare their performance with other teams. Compared to the check phase, more extensive datasets were used, drawn from private sources that either matched or closely resembled the distribution of the final test sets. None of the validation benchmark data was publicly available.

Each task had its own individual validation leaderboard, supporting users in evaluating performance in isolation. To encourage the development of generalizable solutions, combined leaderboards were available, aggregating performance across related tasks based on domain and modality. Four combined leaderboard types were defined: pathology-vision tasks, radiology-vision tasks, language tasks, and an \textit{all-tasks} leaderboard. The all-tasks leaderboard spanned all 20 tasks of the challenge and served as the global benchmark.

To balance evaluation frequency with computational constraints, each team was allowed a maximum of six successful validation submissions per task: three per task-specific leaderboard, two per domain/modality-based combined leaderboard, and one to the all-tasks leaderboard. All successful validation submissions were displayed on their respective leaderboards, allowing users to benchmark their progress and use this feedback to guide algorithm development ahead of the final test phase.

\subsection{Test phase}
During the test phase, algorithms were evaluated on sequestered test sets. The primary goal of the challenge was to benchmark the generalizability of foundation models across diverse tasks, rather than optimizing for individual tasks in isolation. Therefore, only combined leaderboards were set up, as described in \ref*{app:challenge_phase_validation}.

Teams could submit to a test leaderboard only once, and could submit either to the all-tasks or to the combined leaderboards, but not both. If they chose to submit to the all-tasks, their results were automatically considered for ranking within the corresponding combined leaderboards when determining challenge winners.

Overall model performance was summarized via the ``UNICORN Score'' introduced in Section \ref{sec:UNICORN_score}.

Final rankings were based on aggregated normalized performance metrics across all tasks in each respective combined leaderboard, providing a comprehensive assessment of each model’s robustness and adaptability. For the overall test leaderboard, the final ranking score is given by Equation \ref{eq:unicorn_score_all}.

\begin{equation}
\label{app_eq:unicorn_score}
\begin{split}
    S_{UNICORN,all} &= \frac{1}{20} \sum_{i=1}^{20} \frac{S_i-S_{majority}}{1-S_{majority}}\\
    &=\frac{1}{20}(S_1+\frac{S_2-0.5}{0.5}+\frac{S_3-0.5}{0.5}+S_4+S_5+\frac{S_6-0.25}{0.75}+S_7+S_8\\
    &\quad+\frac{S_9-0.2548}{0.7452} + S_{10}+S_{11}+S_{12}+\frac{S_{13}-0.5}{0.5}+\frac{S_{14}-0.5}{0.5}+S_{15}\\
    &\quad+\frac{S_{16}-0.5}{0.5}+\frac{S_{17}-0.7580}{0.2420}+\frac{S_{18}-0.7668}{0.2332}+S_{19}+S_{20})
\end{split}
\end{equation}

\section{Example code}
\label{app:example_code}
To help users get started, a set of open-source repositories covering various aspects of the challenge was released. These resources serve both as functional templates and as reference implementations to support local development and on-platform submissions.

The \emph{minimal working template}\footnote{\url{https://github.com/DIAGNijmegen/unicorn_baseline_template}} includes only the essential code for preprocessing inputs and formatting outputs according to the platform’s specifications. This template does not perform actual model inference; instead, it uses intensity statistics as stand-in features, allowing users to verify integration and understand the expected input/output structure.

The \emph{full reference solution} provides a baseline solution\footnote{\url{https://github.com/DIAGNijmegen/unicorn_baseline}} using publicly available models. The repository includes end-to-end logic for all supported tasks, including data preprocessing and feature extraction. This implementation, based on publicly available models and equipped with lightweight adaptors, successfully executes across all tasks and produces a UNICORN score of 0.378.

Finally, a repository with \emph{adaptor strategies and evaluation logic}\footnote{\url{https://github.com/DIAGNijmegen/unicorn_eval}} contains adaptor implementations for the vision tasks and the evaluation code, making all task-specific metric computations publicly available. Benchmark users were encouraged to contribute new adaptors via pull requests.

\section{Tables}
\begin{table}[!h]
\centering
\caption{Public Datasets for the Development Phase}
\label{app_tab:dataset_info}
\begin{adjustwidth}{-1cm}{-1cm}
\begin{tabular}[c]{|l|l|p{3cm}|p{8cm}|l|}
\hline
\textbf{Task} & \textbf{Dataset} & \textbf{License} & \textbf{Public Links} & \textbf{Publication} \\ \hline
1  & PANDA & CC BY-NC-SA 4.0 & \url{https://www.kaggle.com/c/prostate-cancer-grade-assessment/data} & \cite{bulten2022artificial} \\ \hline
2 & LUNA25 & CC-BY \newline CC-BY-NC & \url{https://zenodo.org/records/14223624}\newline \url{https://zenodo.org/records/14673658} & \cite{venkadesh2021deep} \\ \hline
3 & LEOPARD & CC BY-NC-SA & \url{https://leopard.grand-challenge.org/} & \cite{leopard} \\ \hline
4  & LUNG\_18-193 & CC BY 4.0 & \url{https://www.synapse.org/Synapse:syn26722626} & \cite{vanguri2022multimodal} \\ \hline
5  & DigestPath & CC BY 4.0  & \url{https://digestpath2019.grand-challenge.org/} & \cite{da2022digestpath}  \\ \hline
6  & PI-CAI & CC BY 4.0 & \url{https://zenodo.org/record/6624726} & \cite{saha2024artificial} \\ \hline
7 & LUNA16 & CC BY 4.0 & \url{https://zenodo.org/records/3723295} \newline \url{https://zenodo.org/records/4121926} & \cite{SETIO2017} \cite{Armato2011LIDC} \\ \hline
8 &  MIDOG & CC BY 4.0 & \url{https://zenodo.org/records/4643381} & \cite{bertram2019large}  \\ \hline
9 &  TIGER & CC BY-NC 4.0 & \url{https://zenodo.org/records/6014422} \newline \url{https://doi.org/10.5281/zenodo.6014420} & \cite{tiger} \\ \hline
10 & ULS23 & CC BY-NC-SA 4.0   & \url{https://zenodo.org/records/10035161}  \newline \url{https://zenodo.org/records/10050960} \newline \url{https://zenodo.org/records/10054306} \newline \url{https://zenodo.org/records/10054702} \newline \url{https://zenodo.org/records/10057471} \newline \url{https://zenodo.org/records/10056235} & \cite{uls} \\ \hline
11 & SPIDER & CC BY 4.0 & \url{https://zenodo.org/records/10159290} & \cite{van2024lumbar} \\ \hline
12  & DRAGON & CC BY-NC-SA 4.0 & Task 13 of DRAGON: \url{https://github.com/DIAGNijmegen/dragon_sample_reports/}  & \cite{bosma2025dragon} \\ \hline
13 & DRAGON & CC BY-NC-SA 4.0 & Task 2 of DRAGON: \url{https://github.com/DIAGNijmegen/dragon_sample_reports/} & \cite{bosma2025dragon} \\ \hline
14 & DRAGON & CC BY-NC-SA 4.0 & Task 3 of DRAGON: \url{https://github.com/DIAGNijmegen/dragon_sample_reports/} & \cite{bosma2025dragon} \\ \hline
15 & DRAGON & CC BY-NC-SA 4.0 & Task 18 of DRAGON: \url{https://github.com/DIAGNijmegen/dragon_sample_reports/} & \cite{bosma2025dragon} \\ \hline
16 & DRAGON & CC BY-NC-SA 4.0 & Task 15 of DRAGON: \url{https://github.com/DIAGNijmegen/dragon_sample_reports/} & \cite{bosma2025dragon} \\ \hline
17 & DRAGON & CC BY-NC-SA 4.0 & Tasks 22, 23, 24 of DRAGON: \url{https://github.com/DIAGNijmegen/dragon_sample_reports/} & \cite{bosma2025dragon} \\ \hline
18 & DRAGON & CC BY-NC-SA 4.0 & Tasks 19, 20, 21 of DRAGON: \url{https://github.com/DIAGNijmegen/dragon_sample_reports/} & \cite{bosma2025dragon} \\ \hline
19 & DRAGON & CC BY-NC-SA 4.0 & Task 25 of DRAGON: \url{https://github.com/DIAGNijmegen/dragon_sample_reports/} & \cite{bosma2025dragon} \\ \hline
20 & WsiCaption & N/A & \url{https://github.com/cpystan/Wsi-Caption} & \cite{Che_WsiCaption_MICCAI2024} \\ \hline
\end{tabular}
\end{adjustwidth}
\end{table}

\printbibliography
\end{refsection}

\end{document}